\pdfoutput=1
\documentclass[10pt,twocolumn,letterpaper]{article}

\usepackage[pagenumbers]{cvpr} 

\usepackage{graphicx}
\usepackage{amsmath}
\usepackage{amssymb}
\usepackage{booktabs}
\usepackage{pifont}
\usepackage{cuted}
\usepackage{capt-of}
\usepackage{bbding}
\usepackage{bm}
\usepackage[etex=true,export]{adjustbox}
\usepackage{float}
\usepackage{caption, subcaption, multirow, overpic, textpos}
\usepackage[table]{xcolor}
\usepackage{physics}

\definecolor{citecolor}{HTML}{0071BC}
\definecolor{linkcolor}{HTML}{ED1C24}
\usepackage[pagebackref,breaklinks,colorlinks,citecolor=citecolor,linkcolor=linkcolor]{hyperref}

\definecolor{my_blue}{HTML}{0050F0}
\definecolor{my_purple}{HTML}{7030A0}
\definecolor{my_light}{HTML}{CAB2D6}
\definecolor{my_green}{HTML}{00C050}

\def\cvprPaperID{0000} 
\def\confName{CVPR}
\def\confYear{2023}

\makeatletter
\newcommand*\bigcdot{\mathpalette\bigcdot@{.5}}
\newcommand*\bigcdot@[2]{\mathbin{\vcenter{\hbox{\scalebox{#2}{$\m@th#1\bullet$}}}}}
\makeatother

\usepackage[capitalize]{cleveref}
\crefname{section}{Sec.}{Secs.}
\Crefname{section}{Section}{Sections}
\Crefname{table}{Table}{Tables}
\crefname{table}{Tab.}{Tabs.}

\newlength\savewidth

\renewcommand{\paragraph}[1]{\vspace{1.25mm}\noindent\textbf{#1}}

\newcolumntype{x}[1]{>{\centering\arraybackslash}p{#1pt}}
\newcolumntype{y}[1]{>{\raggedright\arraybackslash}p{#1pt}}
\newcolumntype{z}[1]{>{\raggedleft\arraybackslash}p{#1pt}}

\newcommand{\app}{\raise.17ex\hbox{$\scriptstyle\sim$}}

\definecolor{deemph}{gray}{0.6}

\definecolor{baselinecolor}{gray}{.9}

\definecolor{my_red}{HTML}{FE4444}
\definecolor{Highlight}{HTML}{39b54a}  
\definecolor{Gray}{gray}{0.95}

\begin{document}

\title{Ref-NPR: Reference-Based Non-Photorealistic Radiance Fields for \\
Controllable Scene Stylization}

\author{Yuechen Zhang$^{1,2}$\hspace{1.0cm}Zexin He$^{1}$\hspace{1.0cm}Jinbo Xing$^{1}$\hspace{1.0cm}Xufeng Yao$^{1}$\hspace{1.0cm}Jiaya Jia$^{1,2}$\\
$^{1}$The Chinese University of Hong Kong~~~
$^{2}$SmartMore\\
\small
\texttt{\{yczhang21, zxhe22, jbxing, xfyao, leojia\}@cse.cuhk.edu.hk}
\vspace{-10pt}
}

\maketitle
\tolerance=1000

\begin{strip}\centering
\vspace{-30pt}
\captionsetup{type=figure}
\includegraphics[width=1\textwidth]{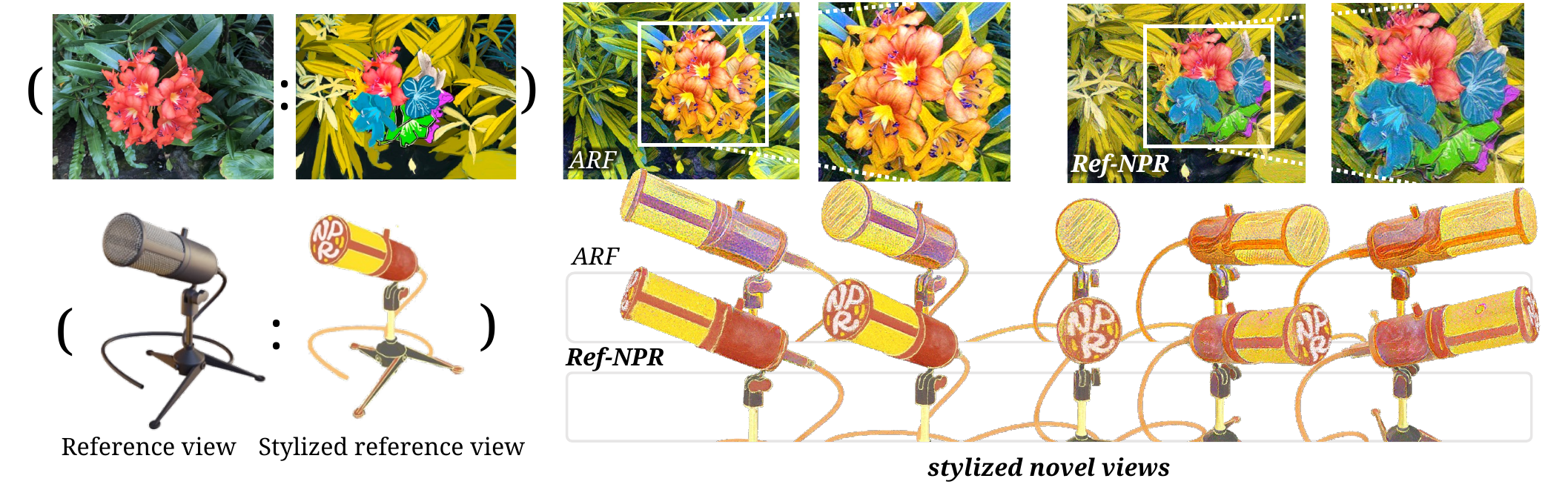}
\vspace{-13pt}
\captionof{figure}{
    Given a pair of one reference view from a radiance field and its stylization, Ref-NPR propagates style more faithfully to novel views with semantic correspondence compared with state-of-the-art scene stylization method ARF~\cite{zhang2022arf}.}
    \label{fig:teaser_image}
\end{strip}

\begin{abstract}
\vspace{-4pt}
Current 3D scene stylization methods transfer textures and colors as styles using arbitrary style references, lacking meaningful semantic correspondences. We introduce Reference-Based Non-Photorealistic Radiance Fields (Ref-NPR) to address this limitation. This controllable method stylizes a 3D scene using radiance fields with a single stylized 2D view as a reference. We propose a ray registration process based on the stylized reference view to obtain pseudo-ray supervision in novel views. Then we exploit semantic correspondences in content images to fill occluded regions with perceptually similar styles, resulting in non-photorealistic and continuous novel view sequences. Our experimental results demonstrate that Ref-NPR outperforms existing scene and video stylization methods regarding visual quality and semantic correspondence. The code and data are publicly available on the project page at \url{https://ref-npr.github.io}.
\end{abstract}

\vspace{-5mm}
\section{Introduction}
\label{sec:intro}

In the past decade, there has been a rising demand for stylizing and editing 3D scenes and objects in various fields, including augmented reality, game scene design, and digital artwork. Traditionally, professionals achieve these tasks by creating 2D reference images and converting them into stylized 3D textures. However, establishing direct cross-modal correspondence is challenging and often requires significant time and effort to obtain stylized texture results similar to the 2D reference schematics.

A critical challenge in the 3D stylization problem is to ensure that stylized results are perceptually similar to the given style reference. Benefiting from radiance fields~\cite{mildenhall2020nerf, kaizhang2020nerfpp, mueller2022instant, chen2022tensorf, yu2022plenoxels, Tancik_2022blocknerf}, recent novel-view stylization methods~\cite{huang_2021_3d_scene_stylization, chiang2022stylizing, Huang22StylizedNeRF, nguyen2022snerf, zhang2022arf, fan2022unified} greatly facilitated style transfer from an arbitrary 2D style reference to 3D implicit representations. However, these methods do not provide explicit control over the generated results, making it challenging to specify the regions where certain styles should be applied and ensure the visual quality of the results. On the other hand, reference-based video stylization methods allow for the controllable generation of stylized novel views with better semantic correspondence between content and style reference, as demonstrated in works like~\cite{jamrivska2019stylizingexample, Texler20fewshot}. However, these methods may diverge from the desired style when stylizing a frame sequence with unseen content, even with the assistance of stylized keyframes.

To address the aforementioned limitations, we propose a new paradigm for stylizing 3D scenes using a single stylized reference view. Our approach, called Reference-Based Non-Photorealistic Radiance Fields (Ref-NPR), is a controllable scene stylization method that takes advantage of volume rendering to maintain cross-view consistency and establish semantic correspondence in transferring style across the entire scene.

Ref-NPR utilizes stylized views from radiance fields as references instead of arbitrary style images to achieve both flexible controllability and multi-view consistency. A reference-based ray registration process is designed to project the 2D style reference into 3D space by utilizing the depth rendering of the radiance field. This process provides pseudo-ray supervision to maintain geometric and perceptual consistency between stylized novel views and the stylized reference view. To obtain semantic style correspondence in occluded regions, Ref-NPR performs template-based feature matching, which uses high-level semantic features as implicit style supervision. The correspondence in the content domain is utilized to select style features in the given style reference, which are then used to transfer style globally, especially in occluded regions. By doing so, Ref-NPR generates the entire stylized 3D scene from a single stylized reference view.

Ref-NPR produces visually appealing stylized views that maintain both geometric and semantic consistency with the given style reference, as presented in~\cref{fig:teaser_image}. The generated stylized views are perceptually consistent with the reference while also exhibiting high visual quality across various datasets. We have demonstrated that Ref-NPR, when using the same stylized view as reference, outperforms state-of-the-art scene stylization methods~\cite{zhang2022arf, nguyen2022snerf} both qualitatively and quantitatively.

In summary, our paper makes three contributions. Firstly, we introduce a new paradigm for stylizing 3D scenes that allows for greater controllability through the use of a stylized reference view. Secondly, we propose a novel approach called Ref-NPR, consisting of a reference-based ray registration process and a template-based feature matching scheme to achieve geometrically and perceptually consistent stylizations. Finally, our experiments demonstrate that Ref-NPR outperforms state-of-the-art scene stylization methods such as ARF and SNeRF both qualitatively and quantitatively. More comprehensive results and a demo video can be found in the supplementary material and on our project page.
 
\section{Related Works}

\subsection{Stylization in 2D}
\paragraph{Arbitrary style transfer} is a well-studied problem in Non-Photorealistic Rendering (NPR)\cite{gooch2001npr, kyprianidis2012state}. Gatys et al.\cite{gatys2016image} first introduced the idea of representing image style as high-level features extracted from pre-trained deep neural networks. Since then, various parametric image style transfer methods~\cite{karras2019adain, johnson2016perceptual, li2018closed, li2017universal, Chen2016FastPS, kolkin2022nnst} have been developed to generate high-quality stylized images efficiently. Video stylization methods that use arbitrary style input~\cite{deng2021mccnet, chen2017coherent, ruder2018artistic, wang2020rerevst, wu2022ccpl} mainly focus on maintaining temporal coherence to achieve continuous stylized frames. Nonetheless, these stylization methods lack interpretability and controllability, even when using stylized keyframes as reference.

\paragraph{Example-based stylization} methods use multiple style references to stylize images while ensuring semantic correspondences between them. Methods such as~\cite{liao2017analogy, he2019progressive} use multi-level semantic feature matching to establish dense correspondences between the content image and the style reference with similar semantics. To facilitate this feature matching, some methods use explicit alignments such as warping or content-aligned stylizing~\cite{selim2016portraitst, Texler20fewshot, jamrivska2019stylizingexample, shih2014style}. These methods offer controllability by allowing editing of the style reference, but they are not suitable for stylizing novel views as they cannot correctly stylize unseen regions using 2D reference-based methods.

\vspace{-1mm}
\subsection{Stylization in 3D}

\paragraph{Reference-based 3D stylization} methods have achieved promising results without the use of radiance fields, as seen in previous works such as Texture Map~\cite{PatchBasedTextureMapping}, Texture Field~\cite{oechsle2019texturefield}, StyLit~\cite{fivser2016stylit, sykora2019styleblit}, and StyleProp~\cite{hauptfleisch2020styleprop}. However, these methods have their limitations. For instance, Texture Field is trained on a restricted ShapeNet dataset, limiting the stylization of 3D objects within the trained categories. StyLit, on the other hand, treats each view of a 3D object as multi-channel 2D guidance and applies texture mapping, resulting in flickering artifacts due to a lack of geometric prior. Although StyleProp utilizes multi-view correspondence maps to obtain stylized novel views, it can only work on viewing directions around the reference view.

\paragraph{Stylizing radiance fields} has recently emerged as a popular topic in computer vision research, driven by the growing popularity of radiance fields~\cite{mildenhall2020nerf,chiang2022stylizing,fan2022unified}. Huang et al.\cite{huang_2021_3d_scene_stylization} pioneered the application of scene stylization on implicit 3D representations, followed by StylizedNeRF\cite{Huang22StylizedNeRF}, CLIP-NeRF~\cite{wang2022clip}, NeRF-Art~\cite{wang2022nerf}, INS~\cite{fan2022unified}, SNeRF~\cite{nguyen2022snerf}, and ARF~\cite{zhang2022arf}, which focus on various degrees, like text-driven stylization in CLIP feature space, stylization using unified representation, memory efficiency, and stylization quality. However, these methods lack explicit correspondence modeling, leading to uncontrollable stylized novel views that differ perceptually from the style reference. To address this issue, we propose Ref-NPR. This reference-guided controllable scene stylization method generates stylized novel views with geometric and semantic consistency with a given stylized view reference.

\begin{figure*}[ht]
    \centering
     \includegraphics[width=0.99\linewidth]{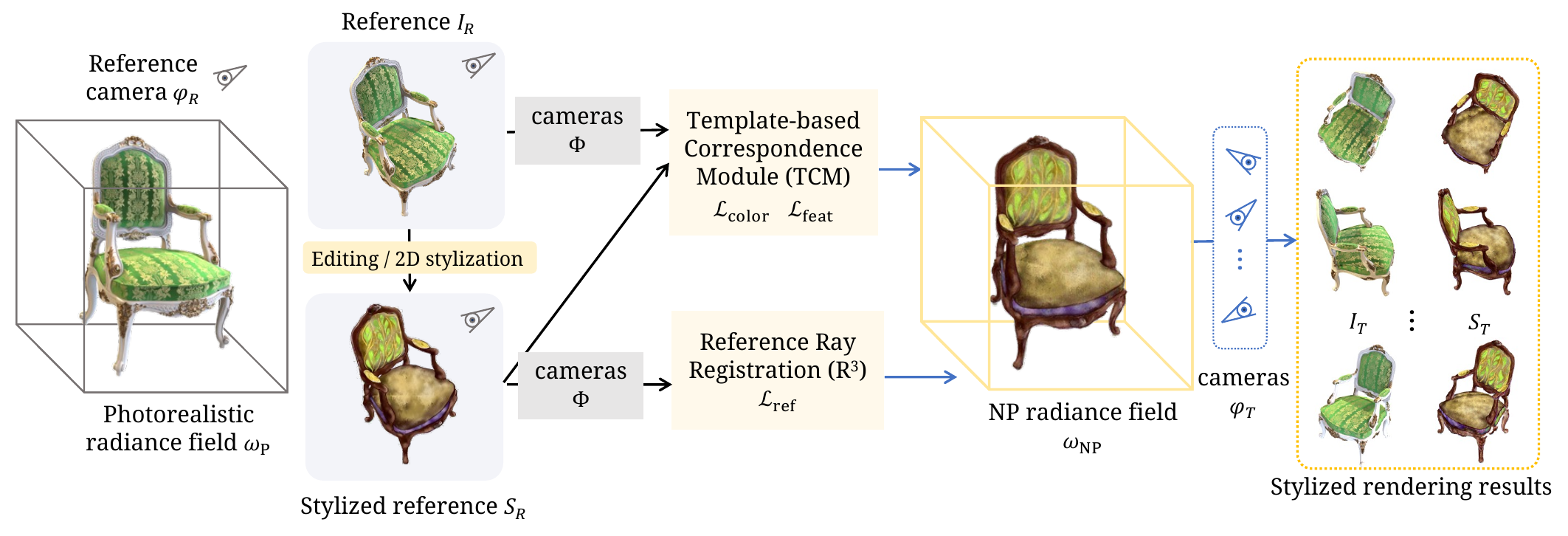}
     \caption{The workflow of \textbf{Ref-NPR}. Given a pre-trained photorealistic radiance field $\omega_{\mathrm{P}}$, we can provide a stylized reference view $S_{R}$ to obtain reference-based supervisions $\mathcal{L}_{\mathrm{ref}}$, $\mathcal{L}_{\mathrm{feat}}$, and $\mathcal{L}_{\mathrm{color}}$. Those loss constraints are used to optimize a non-photorealistic~(NP) radiance field $\omega_{\mathrm{NP}}$. During inference, with such NP radiance field, stylized results $S_{T}$ could be rendered from an arbitrary set of camera poses $\varphi_{T}$, corresponding to the original views $I_{T}$ rendered by $\omega_{\mathrm{P}}$.}
     \label{fig:overview}
     \vspace{-3mm}
 \end{figure*}
 
\section{Ref-NPR}
\cref{fig:overview} outlines the process of Ref-NPR using a single reference view as an example. Ref-NPR aims to stylize a pre-trained photorealistic radiance field with the help of one or a few pairs of reference views and their corresponding stylizations. The first step involves rendering a reference view $I_{R}$ from a specific reference camera $\varphi_{R}$ using a pre-trained photorealistic radiance field $\omega_{\mathrm{P}}$. Next, the reference view $I_{R}$ is stylized by either manually editing or by utilizing structure-preserving 2D-stylization algorithms such as Gatys~\cite{gatys2016image} or AdaIN~\cite{karras2019adain} to obtain a stylized reference view $S_{R}$ based on $I_{R}$.

To enable explicit supervision from the stylized reference view $S_{R}$ to novel views, we propose a Reference Ray Registration (R$^{3}$) process introduced in section \ref{sec:method_rrr}. This process produces a set of reference-dependent pseudo-rays $\Gamma_{R}$, which are correlated rays produced between the reference camera $\varphi_{R}$ and the set of training camera poses $\Phi$. Additionally, we use a Template-based Correspondence Module~(TCM) in section \ref{sec:method_tcm} to obtain implicit style supervision in occluded regions of the reference view. Together, R$^3$ and TCM provide explicit and implicit supervision to optimize a new non-photorealistic (NP) radiance field $\omega_{\mathrm{NP}}$ in section \ref{sec:method_opt}, allowing us to access stylized rendering results of arbitrary target views.

\subsection{Preliminary: Radiance Field Rendering}
\label{sec:method_prelim}
Volume rendering~\cite{kajiya1984ray, mildenhall2020nerf} uses camera rays to sample a 3D radiance field $\omega$ to render images. A camera ray $\mathbf{r}(t)=\mathbf{o} + t\mathbf{d}$ is defined by an origin $\mathbf{o}\in\mathbb{R}^3$ and a direction $\mathbf{d}\in\mathbb{R}^3$ pointing towards the center of a pixel in the image. The radiance field is sampled at $N$ points along the ray, denoted by $\{{\mathbf{r}(t_{i})|i=1\dots N,t_i<t_{i+1}}\}$. At each sample point, the radiance field returns a density value $\sigma_{i} = \sigma(\mathbf{r}(t_i))$ and a view-dependent color $\mathbf{c}_{i} = c(\mathbf{r}(t_{i}), \mathbf{d})$. The accumulated pixel color $\hat{C}(\mathbf{r})$ of the ray $\mathbf{r}$ is estimated in the discrete context, as formulated in~\cite{mildenhall2020nerf}:

\vspace{-3pt}
\begin{equation}
    \hat{C}(\mathbf{r})= \sum_{i=1}^{N} T_{i} (1-\mathrm{exp}(-\sigma_{i}(t_{i+1}-t_{i})))\mathbf{c}_{i} \mathrm{,}
    \label{eq:vol_rendering_color}
\end{equation}
\vspace{-3pt}
\begin{equation}
    \mathrm{where} \,\, T_{i}=\mathrm{exp}\Big(-\sum_{j=1}^{i-1}\sigma_{j}(t_{j+1}-t_{j})\Big),
    \label{eq:vol_rendering_trans}
\end{equation}
which estimates the accumulated transmittance along the ray from $t_1$ to $t_i$.
During training, the radiance field is optimized by directly minimizing the discrepancy between the predicted pixel color $\hat{C}(\mathbf{r})$ and the ground truth pixel color $C(\mathbf{r})$ for each ray, which is denoted by
\vspace{-1pt}
\begin{equation}
    \mathcal{L}_{\omega} = \sum_{\mathbf{r}}\|\hat{C}(\mathbf{r}) - C(\mathbf{r})\|_{2}^{2}.
    \label{eq:vol_rendering_loss}
\end{equation}
\vspace{-3pt}

Once the radiance field is optimized, depth estimation can be performed by mapping the ray $\mathbf{r}$ to an exact 3D position in the scene. This is achieved by setting a threshold $\sigma_z$ on the accumulated density $T_i$ calculated in Equation~\ref{eq:vol_rendering_trans}. The first sample that exceeds this threshold is interpreted as the intersection point between the ray and the scene. The length of the ray $\mathbf{r}$ is then defined as the distance between the camera origin $\mathbf{o}$ and the intersection point, denoted by 

\vspace{-3pt}
\begin{equation}
    l(\mathbf{r}) = \operatorname*{min}\{t_{i} | \sum_{j=1}^{i-1}\sigma_{j}(t_{j+1}-t_{j}) \geq \sigma_{z}\}.
    \label{eq:vol_rendering_depth}
\end{equation}
Then we write the intersection point corresponding to ray $\mathbf{r}$ as $\mathbf{x}({\mathbf{r}}) = \mathbf{o} + l(\mathbf{r})\mathbf{d}$, which is our desired mapping.

\subsection{Reference Ray Registration}
\label{sec:method_rrr}
\begin{figure}
    \centering
    \includegraphics[width=0.95\linewidth]{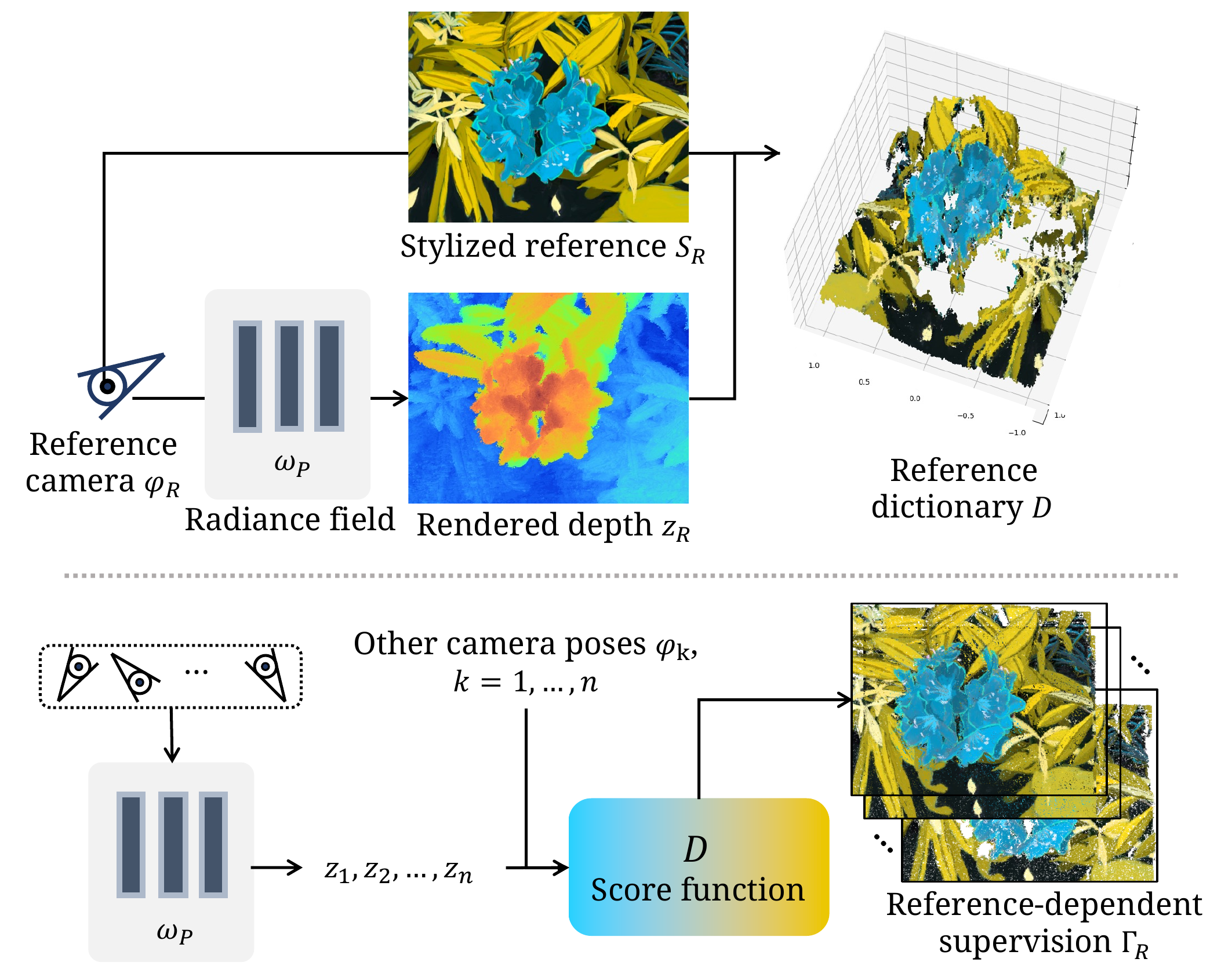}
    \caption{\textbf{Reference Ray Registration.} To create a reference dictionary $D$ for a stylized reference view $S_{R}$ with camera pose $\varphi_{R}$, we estimate pseudo-depths from $\omega_{\mathrm{P}}$. Then, we apply a ray registration process to train camera poses and acquire a collection $\Gamma_{R}$ consisting of pseudo-rays and their assigned colors.}
    \label{fig:rayreg}
    \vspace{-3mm}
\end{figure}

Ref-NPR differs from existing scene stylization methods~\cite{Huang22StylizedNeRF, zhang2022arf, nguyen2022snerf} by incorporating an additional objective of establishing semantic consistency between the stylized reference view and novel views. This objective is also pursued by some scene modeling methods~\cite{kangle2021dsnerf, Xu_2022_SinNeRF, xu2022point}, which use 3D information such as depth to enhance rendering quality. Notably, precise 3D positions and camera parameters enable pixel-wise correspondence between views with a closed-form solution. This is utilized in many methods, such as homographic warping. In Ref-NPR, the radiance field $\omega_{\mathrm{P}}$ is used to estimate pseudo-depth information based on~\cref{eq:vol_rendering_depth}. 

R$^3$, or Reference Ray Registration, utilizes pseudo-depth information to obtain reference-related novel view supervision as shown in~\cref{fig:rayreg}. Using the depth rendering property, pixels in the stylized reference view $S_{R}$ can be mapped to 3D space by estimating their respective ray lengths. A reference dictionary $D$ is constructed, where the element at index $(x, y, z)$ contains all the rays that terminate in the voxel with the same index. This is achieved using a quantization operator $Q(\cdot)$, which maps 3D positions to their corresponding voxels. The reference dictionary is formally defined as
\vspace{-3pt}
\begin{equation}
    D_{(x,y,z)}=\{\mathbf{r}_i \in \varphi_{R}| \, Q(\mathbf{x}(\mathbf{r}_i))=(x,y,z)\},
    \label{eq:regray_dict}
    \vspace{-3pt}
\end{equation}
in which $\varphi_R$ is the reference camera and $\mathbf{x}(\mathbf{r}_i)$ is the intersection point of $\mathbf{r}_i$ estimated from the pseudo-depth $l(\mathbf{r}_i)$ according to~\cref{eq:vol_rendering_depth}. 
By splitting 3D space into discrete voxels, each entry in $D$ may be mapped with multiple rays or none at all, depending on the estimated pseudo-depths.
For each ray $\mathbf{r}_i \in \varphi_{R}$, we use $\hat{C}_R(\cdot)$ to denote the stylized color according to the stylized reference view $S_R$.

Now that $S_R$ is propagated to 3D space with such reference dictionary, we thus register each ray $\mathbf{r}_{j} \in \Phi$ from the training views as a pseudo-ray $\mathbf{\hat{r}}_{j} \in \varphi_R$ in a best-matching manner, through minimizing the Euclidean distance of two corresponding intersection points in 3D space.
Further, to avoid the over-matching problem coming from the gap of ray directions, we deploy a constraint that the angle spanned between directions of matched rays should not exceed a certain threshold $\theta$, i.e., $\angle(\mathbf{d}_{\mathbf{r}_i},\mathbf{d}_{\mathbf{r}_j}) < \theta$. 
We formulate the ray registration process as
\begin{equation}
    \mathbf{\hat{r}}_{j} = \operatorname*{arg\,min}_{\substack{\mathbf{r_{i}} \in D_{(x,y,z)}, \\ \angle(\mathbf{d}_{\mathbf{r}_i},\mathbf{d}_{\mathbf{r}_j}) < \theta}} \|\mathbf{x}(\mathbf{r}_{i})-\mathbf{x}(\mathbf{r}_{j})\|_{2}, 
    \label{eq:regray_query_1}
    \vspace{-3pt}
\end{equation}
\begin{equation}
    \mathrm{where} \,\, Q(\mathbf{x}({\mathbf{r}_j}))=(x,y,z),
    \mathbf{r}_{j} \in \Phi.
    \label{eq:regray_query_3}
\end{equation}

Ray registration finds a ray in the dictionary whose intersection point drops into the same voxel $D_{(x,y,z)}$ as the ray of interest.
Eventually, we construct reference-dependent pseudo-ray supervision as $\Gamma_{R}$ by collecting each validated, registered ray $\mathbf{r}_{j}$ and assign its color in accordance to the corresponding reference ray $\mathbf{\hat{r}}_j$.
Such a collection of pseudo-rays and their stylized colors with the aforementioned $\hat{C}_R(\cdot)$ is defined as
\begin{equation}
    \Gamma_{R} = \{(\mathbf{r}_{j}, \hat{C}_R(\mathbf{\hat{r}}_j)) \,|\, \mathbf{r}_{j} \in \Phi\cup\varphi_R, \mathbf{\hat{r}}_{j}\neq \varnothing\},
    \label{eq:regray_res}
    \vspace{-3pt}
\end{equation}
where $\Phi$ is the collection of all accessible camera poses.

\subsection{Template-Based Semantic Correspondence}
\label{sec:method_tcm}
While R$^3$ is effective in generating pseudo-ray supervision around the given reference camera pose $\varphi_{R}$, it struggles to register reference rays to occluded regions under such a camera, especially for 360$^{\circ}$ scenes in~\cite{Knapitsch2017tnt,mildenhall2020nerf}. To overcome this limitation, we leverage the common assumption in scene stylization that \textit{the semantic correspondence in the entire scene should be consistent before and after stylization}. Based on this, we use the stylized reference view $S_R$ based on source content $I_R$ to establish a content-style mapping as a template. We then broadcast such reference style to novel views with semantically similar content. To achieve this, we introduce a Template-based Correspondence Module~(TCM) that utilizes this content-style correspondence to construct a semantic correlation within the content domain.

\begin{figure}
    \centering
    \includegraphics[width=0.99\linewidth]{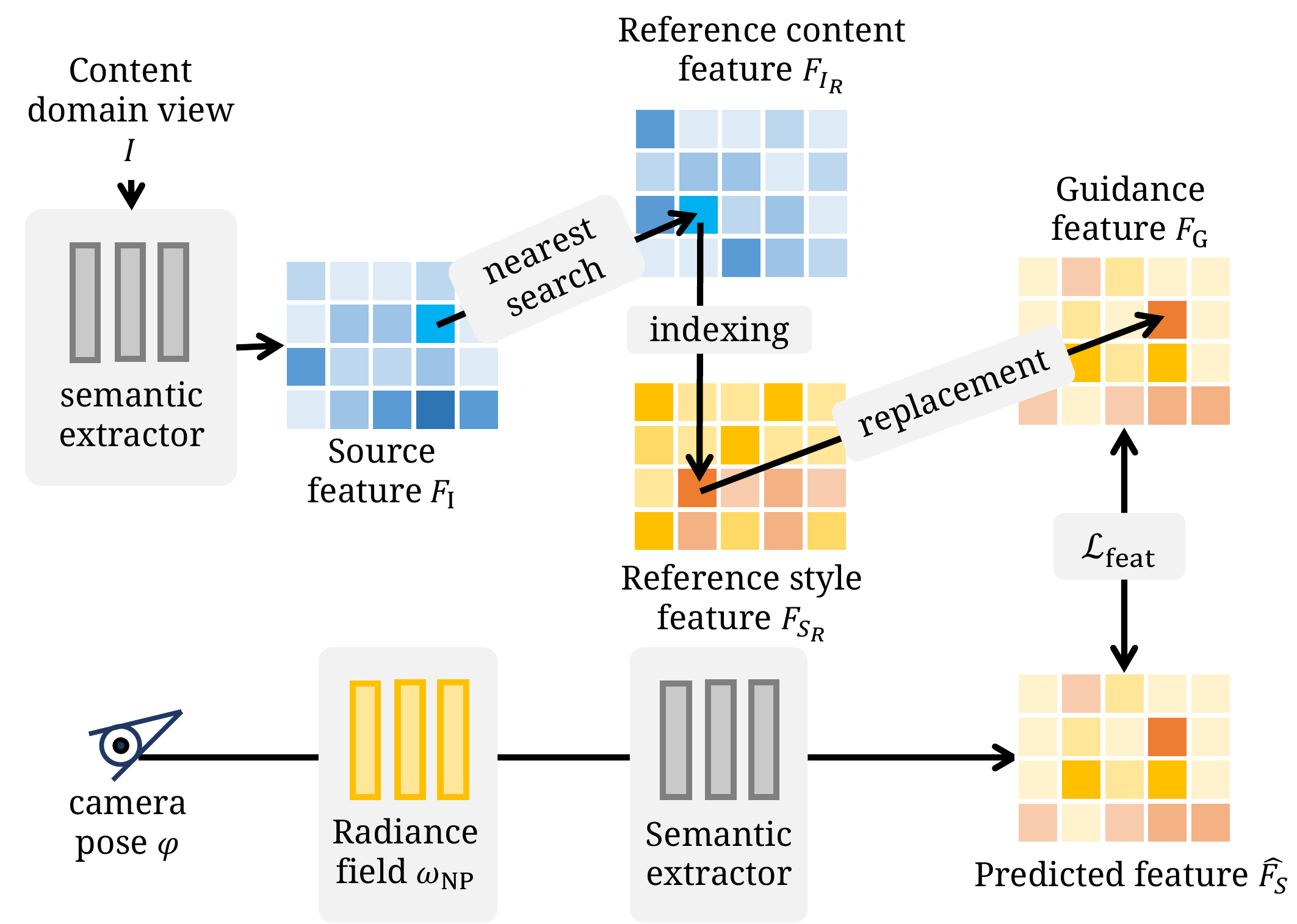}
    \caption{\textbf{Template-based Correspondence Module~(TCM)}. To get the implicit style supervision of one view, the content domain view $I$ is passed to a semantic extractor to obtain its semantic feature $F_{I}$. Then a search-replacement process is conducted to replace the reference feature according to the correspondence in the content domain. The resulting guidance feature $F_{G}$ serves as implicit supervision to optimize the radiance field $\omega_{\mathrm{NP}}$.}
    \label{fig:feat_replace}
    \vspace{-1mm}
\end{figure}

As illustrated in~\cref{fig:feat_replace}, for each content domain view $I$ rendered from $\omega_{\mathrm{P}}$ under some certain camera pose $\varphi \in \Phi$, we obtain its high-level semantic feature map $F_{I}$ from a pre-trained semantic feature extractor~(e.g., VGG16~\cite{Simonyan14VGG}).
Similarly, we denote the extracted feature maps for content reference $I_{R}$ and style reference $S_{R}$ by $F_{I_{R}}$ and $F_{S_{R}}$, respectively, and use a superscript to index each element in the feature maps.
We therefore construct our desired guidance feature $F_{G}$ for further supervision, described by a search-and-replace process on 2D position ${(i, j)}$ as
\vspace{-3pt}
\begin{equation}
    F_{G}^{(i, j)} = F_{S_R}^{(i^{*}, j^{*})},
    \label{eq:tcm_replace}
\end{equation}
\begin{equation}
    \mathrm{where} \,(i^{*}, j^{*}) = \operatorname*{arg\,min}_{i', j'} \, dist\big(F_{I}^{(i, j)}, \, F_{I_R}^{(i', j')}\big).
    \label{eq:tcm_search}
\end{equation}
Here we use $dist(\mathbf{a}, \mathbf{b})$ to denote the distance between two feature vectors $\mathbf{a}$ and $\mathbf{b}$, which is proved to be effective~\cite{zhang2022arf, kolkin2022nnst} by taking the form of cosine distance
when evaluating semantic features.

Finally, for a stylized view $S$ rendered from the NP radiance field $\omega_{\mathrm{NP}}$ under the camera pose $\varphi$, we conduct implicit feature-level supervision by forcing its semantic feature $\hat{F}_{\mathrm{S}}$ to imitate the aforesaid guidance feature ${F}_{G}$.

\subsection{Ref-NPR Optimization}
\label{sec:method_opt}
With the previously discussed collection $\Gamma_{R}$ and guidance feature $F_{G}$ as supervision, we optimize Ref-NPR and obtain the stylized scene representation $\omega_{\mathrm{NP}}$.

In each training iteration, we sample a subset from $\Gamma_{R}$ and denote the set of reference-dependent rays as $N_s$.
For each sampled reference ray $\mathbf{r}_k \in N_s$, $\hat{C}_R(\mathbf{\hat{r}}_k)$ is the assigned color of the corresponding pseudo-ray $\mathbf{\hat{r}}_k$, as defined in~\cref{eq:regray_query_1}, and we further denote $\hat{C}_{\mathrm{NP}}(\mathbf{r}_k)$ to be the stylized color rendered from $\omega_{\mathrm{NP}}$.
This explicit supervision is formulated as the reference loss
\vspace{-3pt}
\begin{equation}
    \mathcal{L}_{\mathrm{ref}} = \frac{1}{|N_s|} \sum_{\mathbf{r}_{k} \in N_s} \| \hat{C}_{\mathrm{NP}}(\mathbf{r}_k) - \hat{C}_R(\mathbf{\hat{r}}_k)\|^2_2 \,.
    \label{eq:opt_ref}
\end{equation}

As for implicit supervision, the discrepancy between $F_{G}$ and $\hat{F}_{\mathrm{S}}$ should be minimized, as discussed in~\cref{sec:method_tcm}.
To maintain the original content structure during such implicit stylization, we also minimize the mean squared distance between content feature $F_{I}$ and stylized feature $\hat{F}_{\mathrm{S}}$, according to~\cite{gatys2016image}.
This implicit supervision is formulated as the feature loss
\vspace{-2mm}
\begin{equation}
\small
    \mathcal{L}_{\mathrm{feat}} = \frac{1}{N} \sum^{N}_{i, j}\big(dist(F_{G}, \hat{F}_{S}) \, + \lambda^{\prime} \| F_I - \hat{F}_{S} \|^{2}_{2} \big),
    \label{eq:opt_feat}
\end{equation}
where $\lambda^{\prime}$ is a balancing factor. 

However, as discussed in~\cite{zhang2022arf, kolkin2022nnst}, optimizing the cosine distance between feature vectors cannot effectively eliminate color mismatches.
To address this issue, we transfer the average color in a patch by a coarse color-matching loss
\begin{equation}
    \mathcal{L}_{\mathrm{color}} = \frac{1}{N}\sum^{N}_{i, j} \| \Bar{C}_{\mathrm{NP}}^{(i, j)} - \Bar{{C}}_{\mathrm{R}}^{(i^*, j^*)} \|^{2}_{2},
    \label{eq:color_loss}
\end{equation}
in which $\Bar{C}_{\mathrm{NP}}^{(i, j)}$ is the $\omega_{\mathrm{NP}}$-rendered average color of the patch at feature-level index $(i, j)$, and $\Bar{{C}}_R^{(i^*, j^*)}$ is the average color of reference patch matched by minimizing feature distance, as described in~\cref{eq:tcm_search}.
Since semantic features are extracted at the image level, considering the memory limitation caused by back-propagation, we follow the gradient cache strategy in~\cite{zhang2022arf} and optimize $\omega_{\mathrm{NP}}$ patch-wisely.

Ultimately, the overall objective is $\mathcal{L}_{\mathrm{NP}} = \lambda_f \mathcal{L}_{\mathrm{feat}} + \lambda_r \mathcal{L}_{\mathrm{ref}} + \lambda_c \mathcal{L}_{\mathrm{color}}$, where $\lambda_{(\cdot)}$ are the balancing factors.
Once $\omega_{\mathrm{NP}}$ is optimized, we may consider it to be a normal radiance field and render stylized novel views with arbitrary camera poses.

\section{Experiments}
\label{sec:exp}
\begin{figure*}
    \centering
     \includegraphics[width=1.0\linewidth]{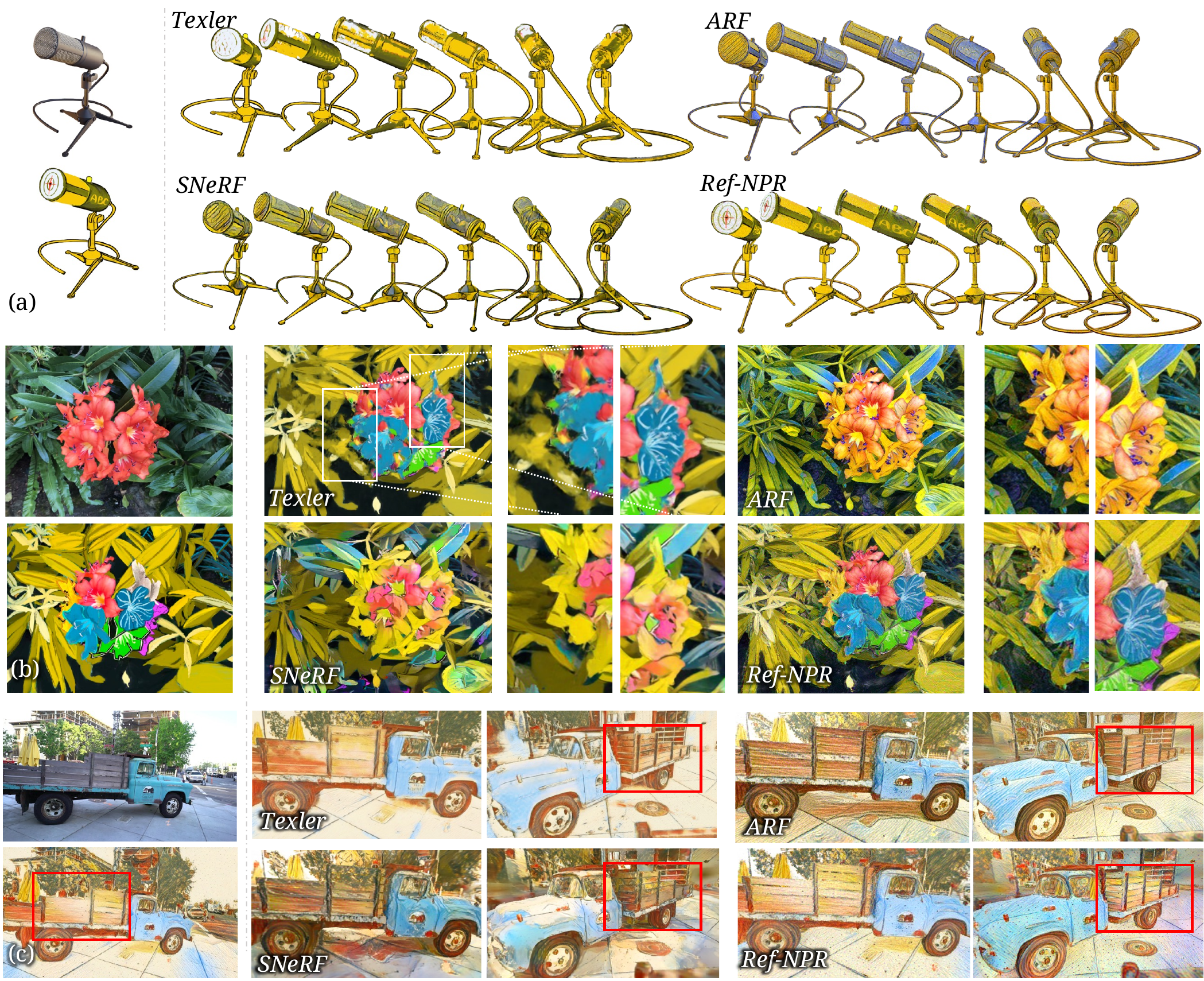}
     \caption{Qualitative comparisons in novel-view stylization. For each example, we provide the reference view~(top) and its corresponding stylized reference view~(bottom) on the left. We compare Ref-NPR with other methods in Synthetic~(a)~\cite{mildenhall2020nerf}, LLFF~(b)~\cite{mildenhall2019llff}, zoomed-in on the flower and the occluded regions, and T\&T~(c)~\cite{Knapitsch2017tnt}. Semantic consistencies are highlighted. }
     \label{fig:compare}
     \vspace{-3mm}
 \end{figure*}

\subsection{Implementation Details}
\label{sec:impl_details}
Ref-NPR is based on the ARF codebase~\cite{zhang2022arf} and uses Plenoxels~\cite{yu2022plenoxels} as the radiance field for scene representation. 
We follow Plenoxels's training scheme to obtain the photorealistic radiance field $\omega_{\mathrm{P}}$.
As we do not expect a view-dependent color change in stylized scenes, following~\cite{Huang22StylizedNeRF, zhang2022arf}, we discard view-dependent rendering and apply a view-independent fitting on training views for two epochs before optimizing $\omega_{\mathrm{NP}}$.
Then, content domain views $I$ are rendered from $\omega_{\mathrm{P}}$ after this view-independent training.
In addition, to keep the same geometrical structure of the scene, we do not optimize the density function $\sigma(\mathbf{r}(t_i))$ in $\omega_{\mathrm{NP}}$~\cite{yu2022plenoxels, zhang2022arf}.

In R$^3$, we set reference dictionary $D$ as a cube containing 256$^3$ voxels.
To parallelize the registration, we store at most 8 rays at each entry $D_{(x, y, z)}$.
The angle constraint of directions in~\cref{eq:regray_query_1} is empirically set to $\cos(\theta) > 0.6$ for ray registration.
For each training step, we set the number of pseudo-ray samples to be $|N_s|=10^{6}$, with half of them from $\varphi_{R}$ and the other half from rays registered in correlated views. 
In TCM, we use VGG16~\cite{Simonyan14VGG} as the semantic feature extractor. We concatenate features that have passed through the activation layers in stages 3 and 4 ($relu\_3\_^{*}$, $relu\_4\_^{*}$), and use them for $\mathcal{L}_{\mathrm{feat}}$.
Balancing factors among loss terms are set to $\lambda^{\prime}=5\times10^{-3}$, $\lambda_{c}=5$, and $\lambda_{r}=\lambda_{f}=1$.
We train each scene on one NVIDIA 3090 GPU for 10 epochs.
Before training the final 3 epochs, for a smoother content update, we replace $\omega_{\mathrm{P}}$ with a frozen $\omega_{\mathrm{NP}}$ as the content view generator in TCM, and minimize $\mathcal{L}_{\mathrm{ref}}$ and $\mathcal{L}_{\mathrm{feat}}$ only, with $\lambda_{f}=0.2$ and $\lambda^{\prime}=0$. 

 \begin{figure*}[ht!]
    \centering
     \includegraphics[width=0.95\linewidth]{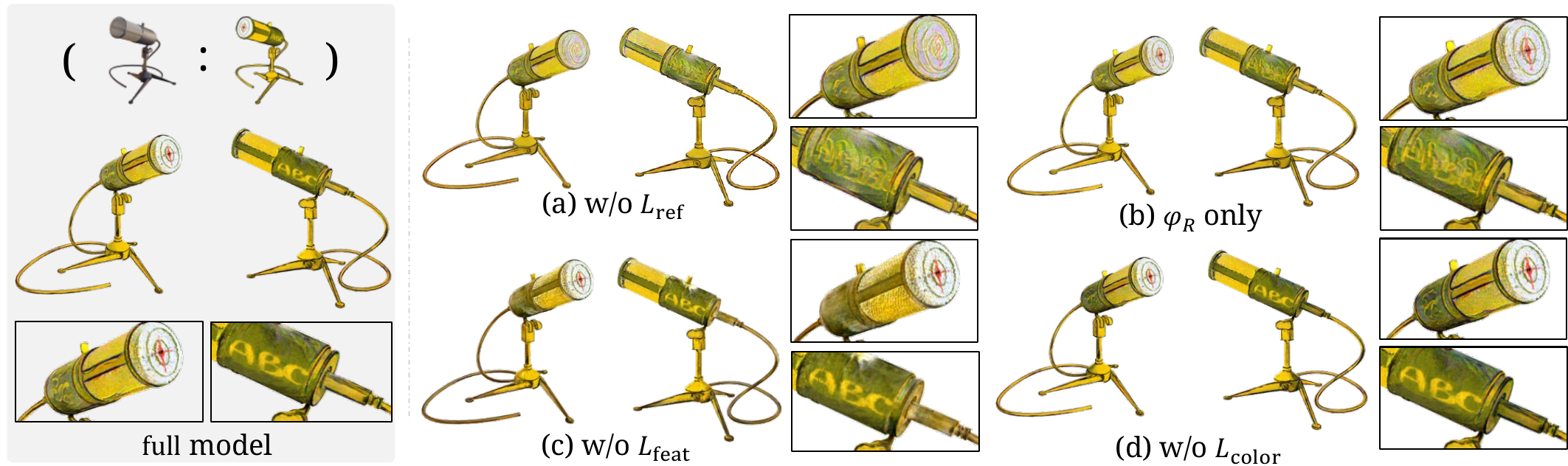}
     \caption{Ablations on the effectiveness of respective loss components.}
     \vspace{-2mm}
     \label{fig:ablation_main_comp}
 \end{figure*}

\subsection{Datasets}
We evaluate the performance of Ref-NPR on three standard datasets.

\paragraph{Synthetic}\cite{mildenhall2020nerf} is a well-defined synthetic dataset for 3D objects. Stylizing the views of 3D models with a foreground mask using a full-image style reference is a challenging yet meaningful task, which is missed in many existing scene stylization methods\cite{huang_2021_3d_scene_stylization, Huang22StylizedNeRF, nguyen2022snerf, zhang2022arf}. To avoid overflow of the stylized view on the mask boundary, we apply a 2D morphological erosion on the foreground mask in R$^3$. 

\paragraph{LLFF}\cite{mildenhall2019llff} is a real-world high-resolution dataset for novel view synthesis, and we choose a resolution that is 4$\times$ downsampled, following\cite{yu2022plenoxels, zhang2022arf}. 

\paragraph{Tanks and Temples~(T\&T)}\cite{Knapitsch2017tnt} is a 360$^{\circ}$ scene dataset for novel view synthesis and scene reconstruction with 200 to 300 high-resolution training views for each scene. In R$^3$, we register only the rays in Plenoxels's foreground model\cite{yu2022plenoxels} apart from the reference view to make the dictionary more compact. We follow the official training and testing splits for all datasets.

\subsection{Comparisons}
\label{sec:comp}
We compare our Ref-NPR with two recent scene stylization methods: ARF~\cite{zhang2022arf} and a reimplemented SNeRF on Plenoxels~\cite{nguyen2022snerf, yu2022plenoxels}.
Besides, we include a reference-based video stylization method by Texler et al.~\cite{Texler20fewshot} for a more comprehensive comparison. 

\paragraph{Qualitative comparison.}
We present qualitative comparisons with state-of-the-art video and scene stylization methods in \cref{fig:compare}. The method of Texler et al.\cite{Texler20fewshot} produces novel views with proper low-level color distribution but suffers from limited correspondence in the scene, leading to flickering effects as shown in \cref{fig:compare}(a). ARF\cite{zhang2022arf} and SNeRF~\cite{nguyen2022snerf} maintain geometric consistency in novel views but lack content-style correlation, whether the style reference is manually created (\cref{fig:compare}(a-b)) or generated by a 2D-neural stylization method~\cite{kolkin2022nnst} in \cref{fig:compare}(c). In contrast, Ref-NPR significantly improves both geometric and semantic-style consistency in each example. Additionally, Ref-NPR benefits from TCM to fill occluded regions with perceptually reasonable visual contents. More detailed comparisons can be found in the supplementary materials. 
 
\paragraph{Quantitative comparison.}
In order to evaluate the quality of reference-based stylization, we adopt a metric used by CCPL~\cite{wu2022ccpl} that measures the consistency between the style reference image $S_R$ and the top 10 nearest test novel views using LPIPS~\cite{zhang2018lpips}. Additionally, we report the long-range cross-view consistency ability as a stability metric, following the approach of SNeRF. We also evaluate the robustness of the methods by running them iteratively and measuring the PSNR between rendered results. Further details on the evaluation metrics are provided in the appendix. The results presented in~\cref{tab:quant_eval} demonstrate that Ref-NPR achieves higher perceptual similarity and cross-view geometric consistency than state-of-the-art scene stylization methods. Moreover, we conduct a user study to evaluate the perceptual quality of the stylization sequences. We apply ten stylization sequences of four shuffled methods and collect 33 responses for each sequence based on the visual quality. The user study results are presented in~\cref{tab:userstudy}.

\begin{table}
	\centering
        \footnotesize
	\renewcommand\arraystretch{1.1}
	{
		\begin{tabular}{y{55}x{27}x{27}x{27}x{35}}
			\toprule
 			metric & Texler & SNeRF & ARF  & \textbf{Ref-NPR} \\
            \midrule
            Stability & \ding{55} & \cellcolor{Gray}\textbf{\ding{51}} & \cellcolor{Gray}\textbf{\ding{51}}  & \cellcolor{Gray}\textbf{\ding{51}}\\
            Ref-LPIPS$~{\textcolor{purple}{\downarrow}}$ & \cellcolor{Gray}\textbf{0.335}  & 0.405 & 0.394 & \cellcolor{Gray}\textbf{0.339}\\
			Robustness$~{\textcolor{purple}{\uparrow}}$  & 18.69 & 26.03 & 26.34 & \cellcolor{Gray}\textbf{28.11}\\
			\bottomrule
		\end{tabular}}
        \caption{Quantitative comparisons. We evaluate the stability, reference similarity, and robustness among stylization methods. Methods except Texler get the same stability as they adopt the same optimized density field.}
        \label{tab:quant_eval}
\end{table}

 \begin{table}
	\centering
        \footnotesize
	\renewcommand\arraystretch{1.1}
	{
		\begin{tabular}{y{55}x{27}x{27}x{27}x{35}}
			\toprule
 			metric & Texler & SNeRF & ARF  & \textbf{Ref-NPR}\\
			\midrule
			Avg. rank$~{\textcolor{purple}{\downarrow}}$ & 2.92 & 2.83 & 2.52  &\cellcolor{Gray}\textbf{1.73}\\
                \textbf{Ours} preference  & 
                \cellcolor{Gray}\textbf{79.3\%}& 
                \cellcolor{Gray}\textbf{78.1\%}& \cellcolor{Gray}\textbf{70.6\%}& - \\
			\bottomrule
		\end{tabular}}
        \caption{User study results. We report the average ranking and pairwise preference rate of Ref-NPR.}
        \vspace{-3mm}
        \label{tab:userstudy}
\end{table}

\subsection{Ablation Studies}
\label{sec:ablation}
We perform several ablation studies to investigate the effectiveness of the different components of Ref-NPR. First, we conduct a module-wise ablation to evaluate the contributions of the supervision components in R$^3$ and TCM. Our results in \cref{fig:ablation_main_comp} (a-d) demonstrate that the removal of any of these components leads to a degradation in the quality of the stylized images. For instance, removing the pseudo-ray supervision $\mathcal{L}_{\mathrm{ref}}$ (a) or supervising on the reference view only (b) results in a loss of detailed textures. While discarding the feature supervision $\mathcal{L}_{\mathrm{feat}}$ (c) causes a photo-realistic colorization in occluded regions. Moreover, neglecting the color supervision $\mathcal{L}_{\mathrm{color}}$ (d) can cause color mismatches in occluded regions. The combination of R$^3$ and TCM in the full Ref-NPR model compensates for these shortcomings and leverages the strengths of each module.

\begin{figure}[t]
    \centering
     \includegraphics[width=0.85\linewidth]{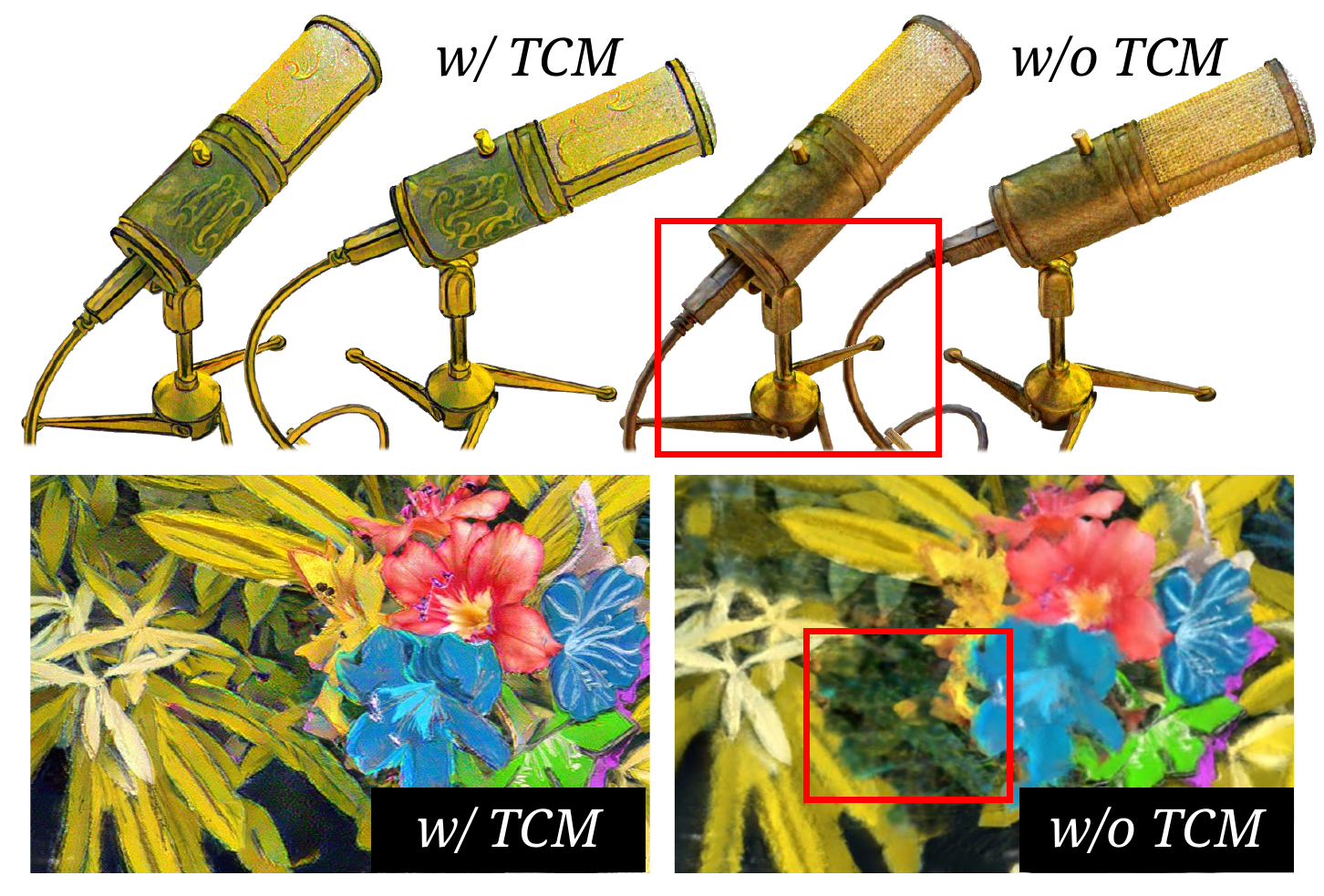}
     \vspace{-1mm}
     \caption{Ablation study on TCM. Benefiting from multi-view correspondence, TCM can effectively fill meaningful styles in occluded regions.}
     \label{fig:ablation_tcm}
     \vspace{-3mm}
 \end{figure}

We also validate the effectiveness of implicit supervision in TCM. 
As depicted in~\cref{fig:ablation_tcm}, we observed that without TCM, the occluded regions often failed to obtain desired correspondence due to the lack of supervision in the semantic feature space. TCM effectively matches features within the same content domain. This demonstrates the crucial role of TCM in achieving accurate and consistent stylization.

\vspace{-1mm}
\section{Discussions}
\label{sec:discuss}
\begin{figure}[H]
    \centering
     \includegraphics[width=1.0\linewidth]{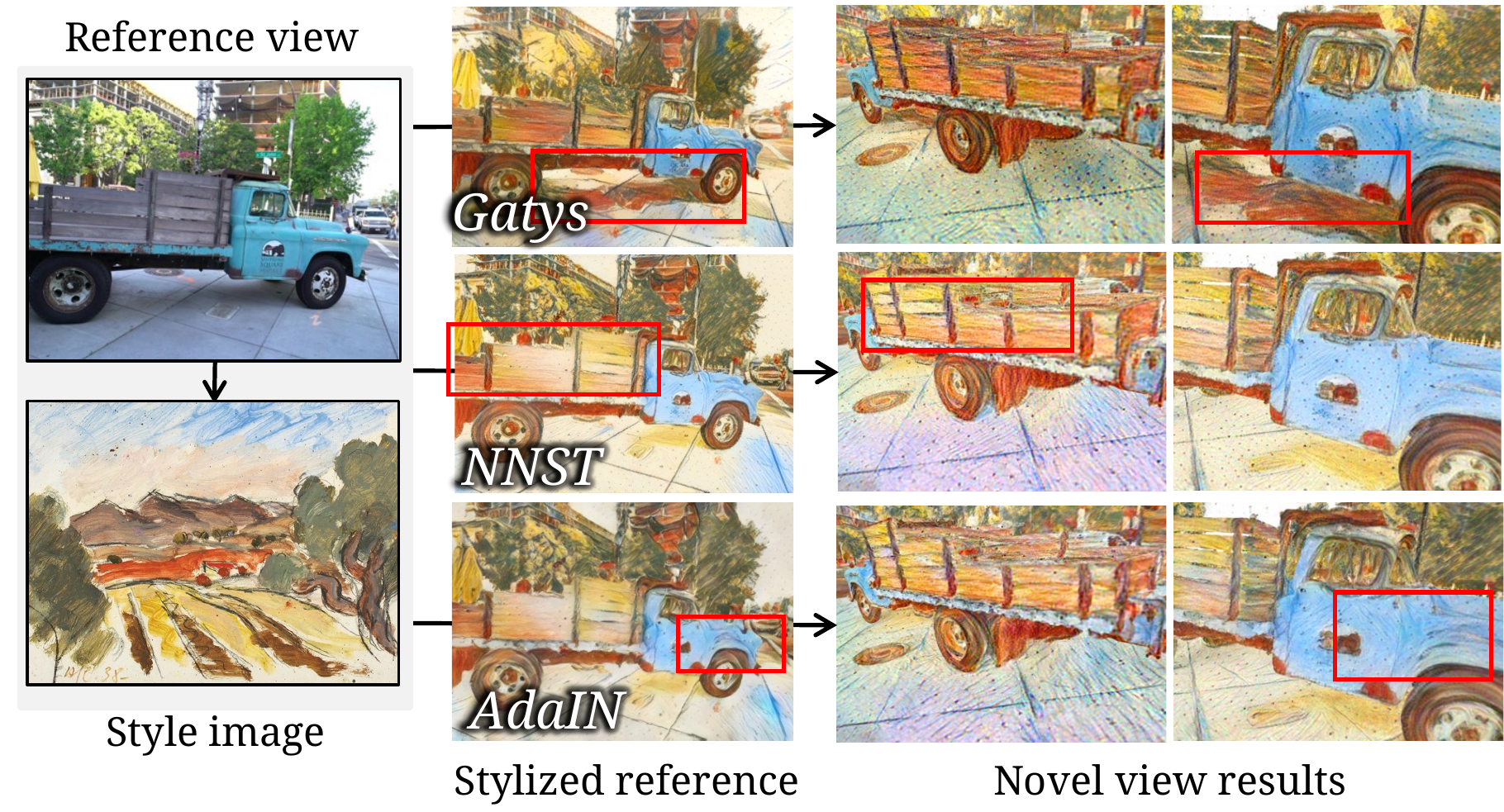}
     \vspace{-3mm}
     \caption{The pipeline of Ref-NPR naturally extends to arbitrary style reference. Images are cropped for a better presentation. Method-related style textures are highlighted.}
     \label{fig:ablation_other_2D_stylize}
     \vspace{-2mm}
\end{figure}

\vspace{-1mm}
\paragraph{Adapt to general style transfer.} 
We have demonstrated the ability of Ref-NPR to use arbitrary style images as reference for style transfer. In particular, as shown in~\cref{fig:ablation_other_2D_stylize}, we generate three reference views of the same scene using different 2D stylization methods~\cite{gatys2016image, karras2019adain, kolkin2022nnst} and feed them into Ref-NPR to render three sets of stylized novel views, each preserving the characteristics of the corresponding style reference. This extension allows Ref-NPR to be more flexible in working with various style reference images and provides a more controllable solution than other scene stylization methods ~\cite{huang_2021_3d_scene_stylization, nguyen2022snerf, Huang22StylizedNeRF, zhang2022arf}.

\begin{figure}[H]
    \centering
     \includegraphics[width=0.9\linewidth]{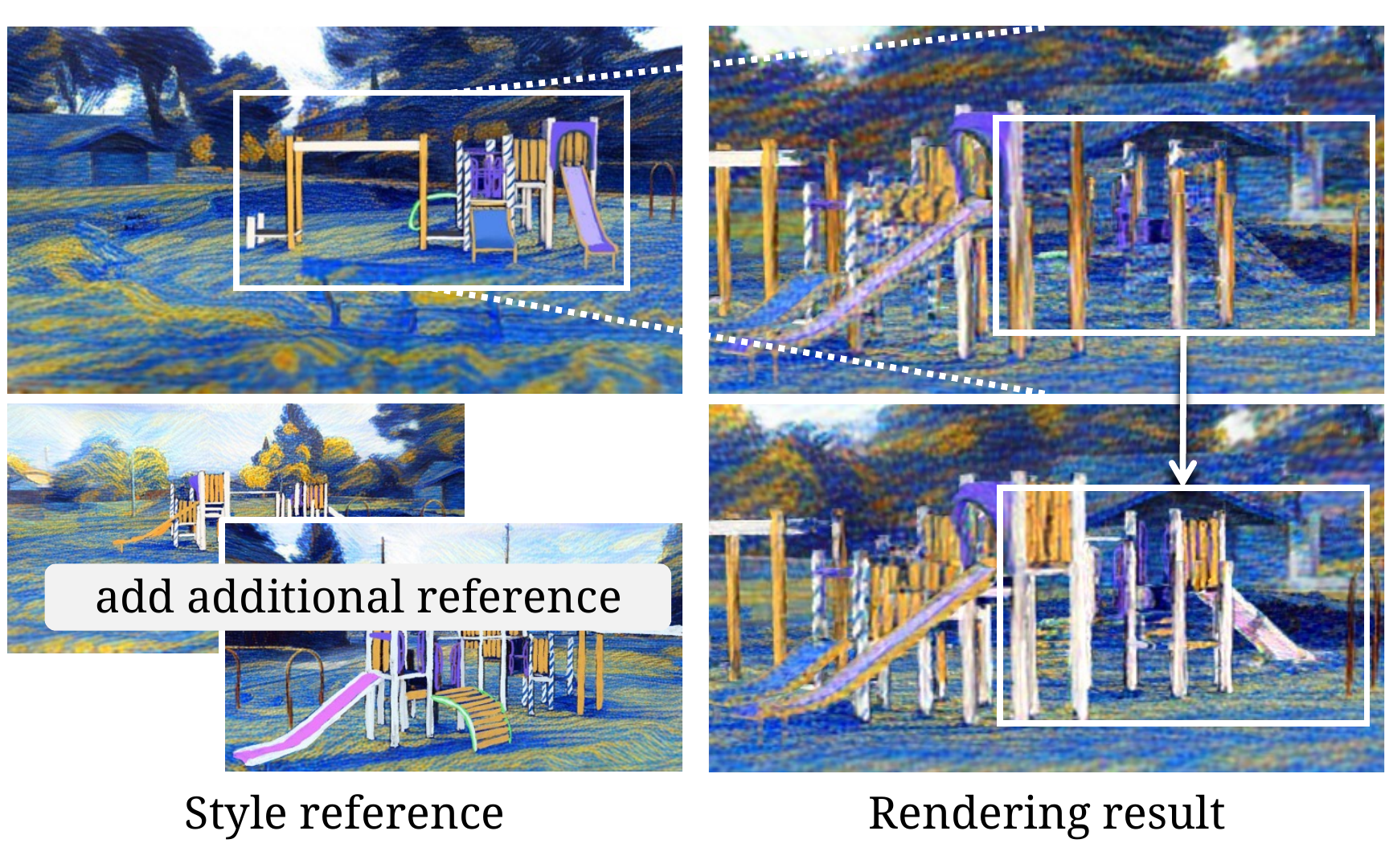}
     \vspace{-3mm}
     \caption{Multi-reference results of a 360$^{\circ}$ scene.}
     \label{fig:ablation_multi_ref}
     \vspace{-5mm}
 \end{figure}

\paragraph{Multi-reference}. 
To accommodate the stylization of large-scale scenes, it is essential to extend Ref-NPR to handle multiple style references. This can be achieved by registering rays using all stylized reference views in R$^3$ and expanding the capacity of styles and content features in TCM. An example of multi-reference input in the Playground scene~\cite{Knapitsch2017tnt} is shown in~\cref{fig:ablation_multi_ref}. Using two additional stylized views, Ref-NPR achieves better feature matching and richer style content.

\vspace{-3mm}

\begin{figure}[H]
    \centering
     \includegraphics[width=0.9\linewidth]{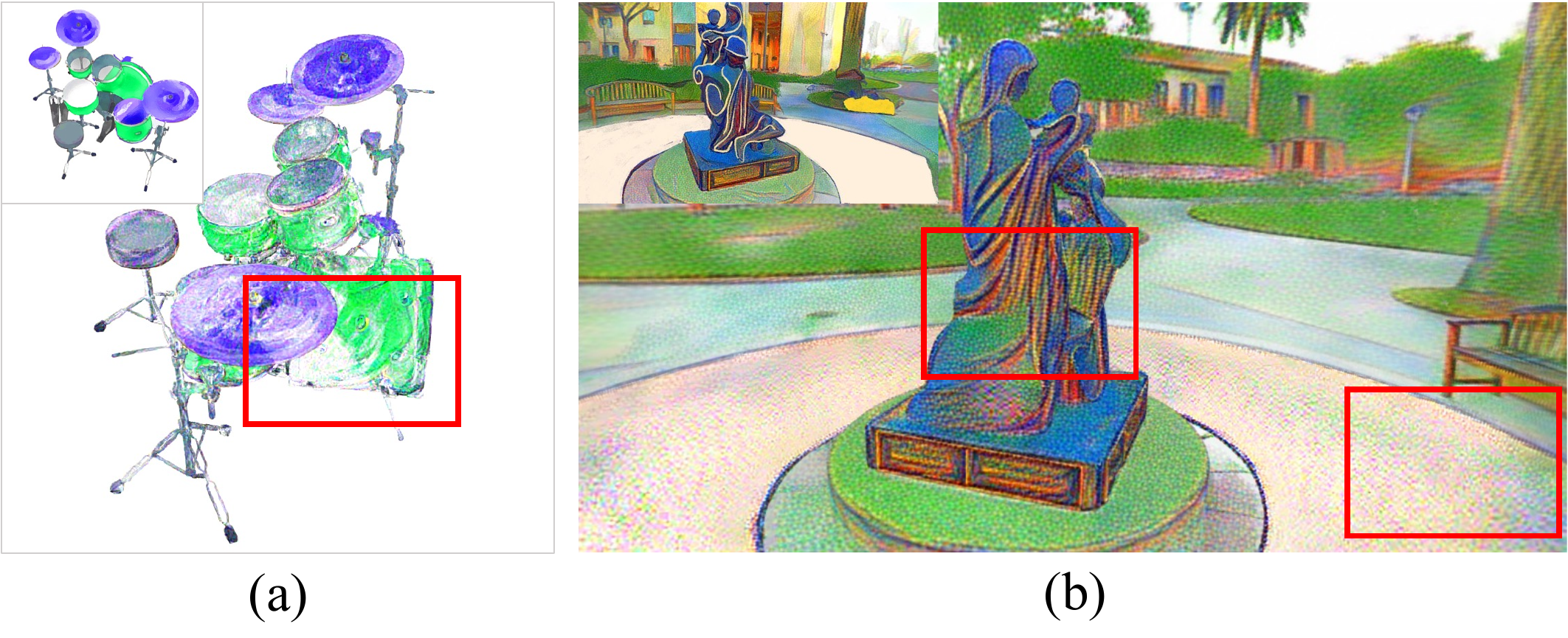}
     \vspace{-1mm}
     \caption{Failure cases of Ref-NPR. Feature matching may fail when (a) stylizing intricate geometric structures, or (b) stylizing a large scene with a single reference.}
     \label{fig:failcase}
     \vspace{-2mm}
\end{figure}

\paragraph{Limitations.} 
Ref-NPR is a versatile method. However, it may not perform well when no meaningful semantic correspondence is found in the reference view, as depicted in Fig~\ref{fig:failcase}. Additionally, feature matching may also fail in stylizing objects with intricate geometric structures.

\section{Conclusion}
\vspace{-1mm}
This paper introduces Ref-NPR, a new framework for controllable non-photorealistic 3D scene stylization based on radiance fields. Ref-NPR can generate high-quality semantic correspondence and geometrically consistent stylizations for novel views by utilizing a stylized reference view. The proposed framework can potentially enhance the efficiency of human creativity in professional visual content creation.

\paragraph{Acknowledgement.} This work is partially supported by ITF Partnership Research Programme (No.PRP/65/20FX) and Shenzhen Science and Technology Program KQTD20- 210811090149095. We are grateful to Shaozuo Yu and Mingxuan Zuo for their meaningful discussions.


{\small
\bibliographystyle{ieee_fullname}
\bibliography{egbib}

\begin{thebibliography}{10}\itemsep=-1pt

\bibitem{PatchBasedTextureMapping}
Sai Bi, Nima~Khademi Kalantari, and Ravi Ramamoorthi.
\newblock Patch-based optimization for image-based texture mapping.
\newblock {\em ACM Trans. Graph.}, 36(4), 2017.

\bibitem{chen2022tensorf}
Anpei Chen, Zexiang Xu, Andreas Geiger, Jingyi Yu, and Hao Su.
\newblock Tensorf: Tensorial radiance fields.
\newblock In {\em ECCV}, 2022.

\bibitem{chen2017coherent}
Dongdong Chen, Jing Liao, Lu Yuan, Nenghai Yu, and Gang Hua.
\newblock Coherent online video style transfer.
\newblock In {\em ICCV}, pages 1105--1114, 2017.

\bibitem{Chen2016FastPS}
Tian~Qi Chen and Mark~W. Schmidt.
\newblock Fast patch-based style transfer of arbitrary style.
\newblock {\em ArXiv}, abs/1612.04337, 2016.

\bibitem{chiang2022stylizing}
Pei-Ze Chiang, Meng-Shiun Tsai, Hung-Yu Tseng, Wei-Sheng Lai, and Wei-Chen
  Chiu.
\newblock Stylizing 3d scene via implicit representation and hypernetwork.
\newblock In {\em CVPR}, pages 1475--1484, 2022.

\bibitem{kangle2021dsnerf}
Kangle Deng, Andrew Liu, Jun-Yan Zhu, and Deva Ramanan.
\newblock Depth-supervised {NeRF}: Fewer views and faster training for free.
\newblock In {\em CVPR}, June 2022.

\bibitem{deng2021mccnet}
Yingying Deng, Fan Tang, Weiming Dong, Haibin Huang, Chongyang Ma, and
  Changsheng Xu.
\newblock Arbitrary video style transfer via multi-channel correlation.
\newblock In {\em AAAI}, volume~35, pages 1210--1217, 2021.

\bibitem{fan2022unified}
Zhiwen Fan, Yifan Jiang, Peihao Wang, Xinyu Gong, Dejia Xu, and Zhangyang Wang.
\newblock Unified implicit neural stylization.
\newblock In {\em ECCV}, pages 636--654. Springer, 2022.

\bibitem{fivser2016stylit}
Jakub Fi{\v{s}}er, Ond{\v{r}}ej Jamri{\v{s}}ka, Michal Luk{\'a}{\v{c}}, Eli
  Shechtman, Paul Asente, Jingwan Lu, and Daniel S{\`y}kora.
\newblock Stylit: illumination-guided example-based stylization of 3d
  renderings.
\newblock {\em ACM Trans. Graph.}, 35(4):1--11, 2016.

\bibitem{yu2022plenoxels}
Sara Fridovich-Keil, Alex Yu, Matthew Tancik, Qinhong Chen, Benjamin Recht, and
  Angjoo Kanazawa.
\newblock Plenoxels: Radiance fields without neural networks.
\newblock In {\em CVPR}, 2022.

\bibitem{gatys2016image}
Leon~A Gatys, Alexander~S Ecker, and Matthias Bethge.
\newblock Image style transfer using convolutional neural networks.
\newblock In {\em CVPR}, pages 2414--2423, 2016.

\bibitem{gooch2001npr}
Bruce Gooch and Amy Gooch.
\newblock {\em Non-photorealistic rendering}.
\newblock AK Peters/CRC Press, 2001.

\bibitem{hauptfleisch2020styleprop}
Filip Hauptfleisch, Ondrej Texler, Aneta Texler, Jaroslav Kriv{\'a}nek, and
  Daniel S{\`y}kora.
\newblock Styleprop: Real-time example-based stylization of 3d models.
\newblock In {\em Computer Graphics Forum}, volume~39, pages 575--586. Wiley
  Online Library, 2020.

\bibitem{he2019progressive}
Mingming He, Jing Liao, Dongdong Chen, Lu Yuan, and Pedro~V Sander.
\newblock Progressive color transfer with dense semantic correspondences.
\newblock {\em ACM Trans. Graph.}, 38(2):1--18, 2019.

\bibitem{huang_2021_3d_scene_stylization}
Hsin-Ping Huang, Hung-Yu Tseng, Saurabh Saini, Maneesh Singh, and Ming-Hsuan
  Yang.
\newblock Learning to stylize novel views.
\newblock In {\em ICCV}, 2021.

\bibitem{Huang22StylizedNeRF}
Yi-Hua Huang, Yue He, Yu-Jie Yuan, Yu-Kun Lai, and Lin Gao.
\newblock Stylizednerf: Consistent 3d scene stylization as stylized nerf via
  2d-3d mutual learning.
\newblock In {\em CVPR}, 2022.

\bibitem{jamrivska2019stylizingexample}
Ond{\v{r}}ej Jamri{\v{s}}ka, {\v{S}}{\'a}rka Sochorov{\'a}, Ond{\v{r}}ej
  Texler, Michal Luk{\'a}{\v{c}}, Jakub Fi{\v{s}}er, Jingwan Lu, Eli Shechtman,
  and Daniel S{\`y}kora.
\newblock Stylizing video by example.
\newblock {\em ACM Trans. Graph.}, 38(4):1--11, 2019.

\bibitem{johnson2016perceptual}
Justin Johnson, Alexandre Alahi, and Li Fei-Fei.
\newblock Perceptual losses for real-time style transfer and super-resolution.
\newblock In {\em ECCV}, pages 694--711. Springer, 2016.

\bibitem{kajiya1984ray}
James~T Kajiya and Brian~P Von~Herzen.
\newblock Ray tracing volume densities.
\newblock {\em ACM Trans. Graph.}, 18(3):165--174, 1984.

\bibitem{karras2019adain}
Tero Karras, Samuli Laine, and Timo Aila.
\newblock A style-based generator architecture for generative adversarial
  networks.
\newblock In {\em CVPR}, pages 4401--4410, 2019.

\bibitem{Knapitsch2017tnt}
Arno Knapitsch, Jaesik Park, Qian-Yi Zhou, and Vladlen Koltun.
\newblock Tanks and temples: Benchmarking large-scale scene reconstruction.
\newblock {\em ACM Trans. Graph.}, 36(4), 2017.

\bibitem{kolkin2022nnst}
Nicholas Kolkin, Michal Kucera, Sylvain Paris, Daniel Sykora, Eli Shechtman,
  and Greg Shakhnarovich.
\newblock Neural neighbor style transfer.
\newblock {\em arXiv e-prints}, pages arXiv--2203, 2022.

\bibitem{kyprianidis2012state}
Jan~Eric Kyprianidis, John Collomosse, Tinghuai Wang, and Tobias Isenberg.
\newblock State of the" art”: A taxonomy of artistic stylization techniques
  for images and video.
\newblock {\em TVCG}, 19(5):866--885, 2012.

\bibitem{li2017universal}
Yijun Li, Chen Fang, Jimei Yang, Zhaowen Wang, Xin Lu, and Ming-Hsuan Yang.
\newblock Universal style transfer via feature transforms.
\newblock {\em NeurIPS}, 30, 2017.

\bibitem{li2018closed}
Yijun Li, Ming-Yu Liu, Xueting Li, Ming-Hsuan Yang, and Jan Kautz.
\newblock A closed-form solution to photorealistic image stylization.
\newblock In {\em ECCV}, pages 453--468, 2018.

\bibitem{liao2017analogy}
Jing Liao, Yuan Yao, Lu Yuan, Gang Hua, and Sing~Bing Kang.
\newblock Visual attribute transfer through deep image analogy.
\newblock {\em ACM Trans. Graph.}, 36(4):120, 2017.

\bibitem{mildenhall2019llff}
Ben Mildenhall, Pratul~P. Srinivasan, Rodrigo Ortiz-Cayon, Nima~Khademi
  Kalantari, Ravi Ramamoorthi, Ren Ng, and Abhishek Kar.
\newblock Local light field fusion: Practical view synthesis with prescriptive
  sampling guidelines.
\newblock {\em ACM Trans. Graph.}, 2019.

\bibitem{mildenhall2020nerf}
Ben Mildenhall, Pratul~P. Srinivasan, Matthew Tancik, Jonathan~T. Barron, Ravi
  Ramamoorthi, and Ren Ng.
\newblock Nerf: Representing scenes as neural radiance fields for view
  synthesis.
\newblock In {\em ECCV}, 2020.

\bibitem{mueller2022instant}
Thomas M\"uller, Alex Evans, Christoph Schied, and Alexander Keller.
\newblock Instant neural graphics primitives with a multiresolution hash
  encoding.
\newblock {\em ACM Trans. Graph.}, 41(4):102:1--102:15, July 2022.

\bibitem{nguyen2022snerf}
Thu Nguyen-Phuoc, Feng Liu, and Lei Xiao.
\newblock Snerf: stylized neural implicit representations for 3d scenes.
\newblock {\em arXiv preprint arXiv:2207.02363}, 2022.

\bibitem{oechsle2019texturefield}
Michael Oechsle, Lars Mescheder, Michael Niemeyer, Thilo Strauss, and Andreas
  Geiger.
\newblock Texture fields: Learning texture representations in function space.
\newblock In {\em Proceedings of the IEEE/CVF International Conference on
  Computer Vision}, pages 4531--4540, 2019.

\bibitem{ruder2018artistic}
Manuel Ruder, Alexey Dosovitskiy, and Thomas Brox.
\newblock Artistic style transfer for videos and spherical images.
\newblock {\em IJCV}, 126(11):1199--1219, 2018.

\bibitem{selim2016portraitst}
Ahmed Selim, Mohamed Elgharib, and Linda Doyle.
\newblock Painting style transfer for head portraits using convolutional neural
  networks.
\newblock {\em ACM Trans. Graph.}, 35(4):1--18, 2016.

\bibitem{shih2014style}
YiChang Shih, Sylvain Paris, Connelly Barnes, William~T Freeman, and Fr{\'e}do
  Durand.
\newblock Style transfer for headshot portraits.
\newblock {\em ACM Trans. Graph.}, 2014.

\bibitem{Simonyan14VGG}
Karen Simonyan and Andrew Zisserman.
\newblock Very deep convolutional networks for large-scale image recognition.
\newblock In Yoshua Bengio and Yann LeCun, editors, {\em ICLR}, 2015.

\bibitem{sykora2019styleblit}
Daniel S{\`y}kora, Ond{\v{r}}ej Jamri{\v{s}}ka, Ondrej Texler, Jakub
  Fi{\v{s}}er, Michal Luk{\'a}{\v{c}}, Jingwan Lu, and Eli Shechtman.
\newblock Styleblit: Fast example-based stylization with local guidance.
\newblock In {\em Computer Graphics Forum}, volume~38, pages 83--91. Wiley
  Online Library, 2019.

\bibitem{Tancik_2022blocknerf}
Matthew Tancik, Vincent Casser, Xinchen Yan, Sabeek Pradhan, Ben Mildenhall,
  Pratul~P. Srinivasan, Jonathan~T. Barron, and Henrik Kretzschmar.
\newblock Block-nerf: Scalable large scene neural view synthesis.
\newblock In {\em CVPR}, pages 8248--8258, June 2022.

\bibitem{Texler20fewshot}
Ond\v{r}ej Texler, David Futschik, Michal Ku\v{c}era, Ond\v{r}ej Jamri\v{s}ka,
  \v{S}\'{a}rka Sochorov\'{a}, Menglei Chai, Sergey Tulyakov, and Daniel
  S\'{y}kora.
\newblock Interactive video stylization using few-shot patch-based training.
\newblock {\em ACM Trans. Graph.}, 39(4):73, 2020.

\bibitem{wang2022clip}
Can Wang, Menglei Chai, Mingming He, Dongdong Chen, and Jing Liao.
\newblock Clip-nerf: Text-and-image driven manipulation of neural radiance
  fields.
\newblock In {\em CVPR}, pages 3835--3844, 2022.

\bibitem{wang2022nerf}
Can Wang, Ruixiang Jiang, Menglei Chai, Mingming He, Dongdong Chen, and Jing
  Liao.
\newblock Nerf-art: Text-driven neural radiance fields stylization.
\newblock {\em arXiv preprint arXiv:2212.08070}, 2022.

\bibitem{wang2020rerevst}
Wenjing Wang, Shuai Yang, Jizheng Xu, and Jiaying Liu.
\newblock Consistent video style transfer via relaxation and regularization.
\newblock {\em TIP}, 29:9125--9139, 2020.

\bibitem{wu2022ccpl}
Zijie Wu, Zhen Zhu, Junping Du, and Xiang Bai.
\newblock Ccpl: Contrastive coherence preserving loss for versatile style
  transfer.
\newblock In {\em ECCV}, 2022.

\bibitem{Xu_2022_SinNeRF}
Dejia Xu, Yifan Jiang, Peihao Wang, Zhiwen Fan, Humphrey Shi, and Zhangyang
  Wang.
\newblock Sinnerf: Training neural radiance fields on complex scenes from a
  single image.
\newblock In {\em ECCV}, 2022.

\bibitem{xu2022point}
Qiangeng Xu, Zexiang Xu, Julien Philip, Sai Bi, Zhixin Shu, Kalyan Sunkavalli,
  and Ulrich Neumann.
\newblock Point-nerf: Point-based neural radiance fields.
\newblock In {\em CVPR}, pages 5438--5448, 2022.

\bibitem{zhang2022arf}
Kai Zhang, Nick Kolkin, Sai Bi, Fujun Luan, Zexiang Xu, Eli Shechtman, and Noah
  Snavely.
\newblock Arf: Artistic radiance fields.
\newblock In {\em Computer Vision--ECCV 2022: 17th European Conference, Tel
  Aviv, Israel, October 23--27, 2022, Proceedings, Part XXXI}, pages 717--733.
  Springer, 2022.

\bibitem{kaizhang2020nerfpp}
Kai Zhang, Gernot Riegler, Noah Snavely, and Vladlen Koltun.
\newblock Nerf++: Analyzing and improving neural radiance fields.
\newblock {\em arXiv:2010.07492}, 2020.

\bibitem{zhang2023adding}
Lvmin Zhang and Maneesh Agrawala.
\newblock Adding conditional control to text-to-image diffusion models, 2023.

\bibitem{zhang2018lpips}
Richard Zhang, Phillip Isola, Alexei~A Efros, Eli Shechtman, and Oliver Wang.
\newblock The unreasonable effectiveness of deep features as a perceptual
  metric.
\newblock In {\em CVPR}, 2018.

\end{thebibliography}
}
\clearpage

	\clearpage
	
	\renewcommand\thesection{\Alph{section}}
	\renewcommand\thesubsection{\thesection.\arabic{subsection}}
	\renewcommand\thefigure{\Alph{section}.\arabic{figure}}
	\renewcommand\thetable{\Alph{section}.\arabic{table}} 
	
	\setcounter{section}{0}
	\setcounter{figure}{0}
	\setcounter{table}{0}
	
	\twocolumn[
	\begin{@twocolumnfalse}
		\begin{center}
			\noindent{\Large{\textbf{Ref-NPR: Reference-Based Non-Photorealistic Radiance Fields for \\ 
						\vspace{0.1in}
						Controllable Scene Stylization: Supplementary Material}}}
		\end{center}
		\vspace{0.4in}
	\end{@twocolumnfalse}
	]

\section{Supplementary Materials}

We have prepared supplementary materials, including a document and a video, to provide a more comprehensive understanding of Ref-NPR.
In the document, we discuss the technical details of our implementation in~\cref{sec:tech_details}, and provide visualizations in~\cref{sec:method_vis} to better illustrate the proposed modules in Ref-NPR.
Moreover, we present additional examples and visualizations in~\cref{sec:supp_comp} to demonstrate the performance and controllability of our method.
Furthermore, we have prepared a video that showcases the results and comparisons of Ref-NPR. We also provide live demo examples on the project page.

\noindent 
\textcolor{citecolor}{\textbf{Video link}} \hspace{2.2mm}  \url{https://youtu.be/jnsnrTwVSBw}.\\
\textcolor{citecolor}{\textbf{Project page}} \url{https://ref-npr.github.io}.

\section{Technical Details}
\label{sec:tech_details}

\paragraph{Implementation details.}
Two worth-noting details may affect the visual quality of stylization results when implementing Ref-NPR.
\begin{itemize}
    \item Before computing image-level loss terms ($\mathcal{L}_{\mathrm{color}}$ and $\mathcal{L}_{\mathrm{feat}}$), for LLFF~\cite{mildenhall2019llff} and T\&T~\cite{Knapitsch2017tnt} dataset, we downsample both stylized and content views by 2x to speed up the calculation of patch-wise feature distance. 
    \item Different from the implicit feature loss $\mathcal{L}_{\mathrm{feat}}$, in order to get a high-level semantic color mapping for the color-matching loss $\mathcal{L}_{\mathrm{color}}$, we evaluate distances between features extracted by the last stage (i.e., stage 5) of VGG backbone~\cite{Simonyan14VGG}. Besides, when calculating $\mathcal{L}_{\mathrm{color}}$, we exclude the position of interest $(i, j)$ where the semantic feature is not close enough to any feature in the reference view, to avoid over-matching.
    Such a constraint of the feature distance for valid position $(i, j)$ is formulated as

\begin{equation}
    \min_{i', j'}\, dist(F_{I}^{(i, j)}, F_{I_R}^{(i', j')}) < 0.4 \,.
    \label{eq:constraint}
\end{equation}

\end{itemize}

\paragraph{Details of comparison.}
Our experiments on Texler~\cite{Texler20fewshot} are conducted using their official implementation. As the reference view can be freely chosen, it is possible that continuous views with high-quality temporal coherence do not appear in the test sequence. Therefore, we only use the RGB image sequence as input and follow the default training settings by training each scene for 30,000 iterations. However, it is important to note that Texler's method is unsuitable for videos with large movements and rotations, and we train it on the template view. Despite applying the Gaussian mixture strategy with a dense sample rate, error accumulation still leads to artifacts in the output.

Regarding SNeRF~\cite{nguyen2022snerf}, we re-implement it based on Plenoxels~\cite{yu2022plenoxels} and use Gatys~\cite{gatys2016image} as the stylization method. We train the stylization step for 10 iterations and the entire scene stylization for 10 epochs for each training view.

\paragraph{Quantitative comparison.}
In Sec.~\textcolor{red}{4.3}, we propose a reference-based perceptual similarity metric to evaluate our method. The detailed LPIPS scores for each scene are reported in~\cref{tab:supp_quant_eval}. It is worth noting that the scene-wise LPIPS scores exhibit significant variations. We speculate that these fluctuations may be due to the substantial differences in camera poses between the reference view and all other test views. Additionally, Texler~\cite{Texler20fewshot} achieves slightly better reference-related LPIPS scores. However, it fails to produce satisfactory results when the camera pose diverges significantly from the reference camera $\varphi_{R}$, as demonstrated in~\cref{fig:supp_compare} and the supplementary video.

\begin{figure*}[ht]
    \centering
    \includegraphics[width=0.95\linewidth]{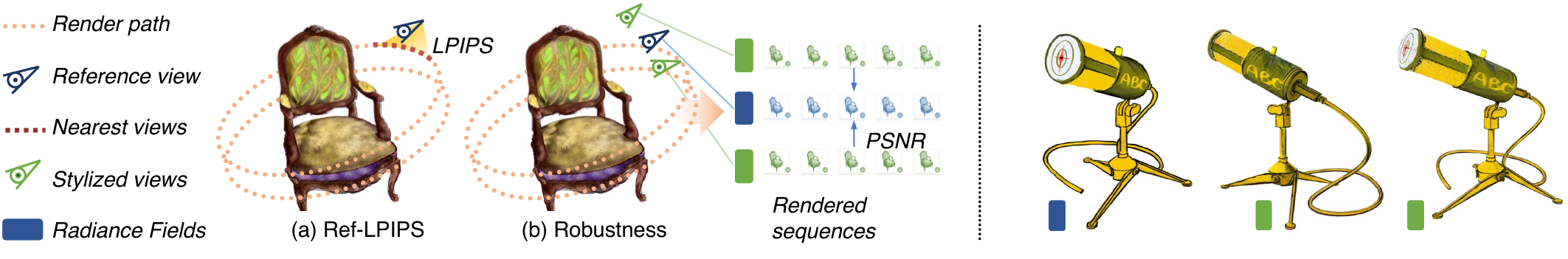}
    \caption{Left: An illustration of the Ref-LPIPS and robustness test. Right: Tested views for robustness.}
    \label{fig:quant_eval}
\end{figure*}

\cref{fig:quant_eval} (a) depicts the procedure of the designed LPIPS evaluation in the paper.
As only the reference image is given to evaluate the visual quality, we utilize LPIPS referring to CCPL~\cite{wu2022ccpl} as a metric for frame-wise stylization consistency. The closest ten frames represent a frame-wise consistency of stylization results with the given style reference.
\cref{fig:quant_eval} (b) depicts our experiments investigating the robustness of stylization methods. For a stylized NeRF $\omega_{\mathrm{NP}}$, we render a set of views as the style reference and use them to get a set of stylized NeRFs. Given the same camera path, we compute the PSNR of rendering results between them and $\omega_{\mathrm{NP}}$.

\begin{table*}
    \centering
    \begin{tabular}{y{55}cccccccccc}
        \hline
         Ref-LPIPS~\textcolor{purple}{$\downarrow$}       & Geo. Consist. & Chair & Ficus & Hotdog & Mic & Flower & Horn & Truck & Playground & Average \\
    
         \hline
         Texler~\cite{Texler20fewshot} & \ding{55} & 0.167 & 0.120 & 0.216 & 0.119 & 0.230 & 0.488 & 0.667 & 0.675 & \cellcolor{Gray}\textbf{0.335}  \\
         ARF~\cite{zhang2022arf}       & \cellcolor{Gray}\ding{51}  & 0.185 & 0.123 & 0.300 & 0.146 & 0.619 & 0.502 & 0.683 & 0.592 & 0.394  \\
         SNeRF~\cite{nguyen2022snerf}  & \cellcolor{Gray}\ding{51} & 0.188 & 0.129 & 0.283 & 0.138 & 0.646 & 0.492 & 0.702 & 0.663 & 0.405  \\
         \textbf{Ref-NPR}              & \cellcolor{Gray}\ding{51} & 0.164 & 0.122 & 0.273 & 0.126 & 0.289 & 0.471 & 0.669 & 0.596 & \cellcolor{Gray}\textbf{0.339}  \\
         \hline
    \end{tabular}
    \caption{Reference-related novel view LPIPS for each test scene.}
    \label{tab:supp_quant_eval}
\end{table*}

\section{Method Visualizations}
\label{sec:method_vis}
\begin{figure}
    \centering
    \includegraphics[width=1.0\linewidth]{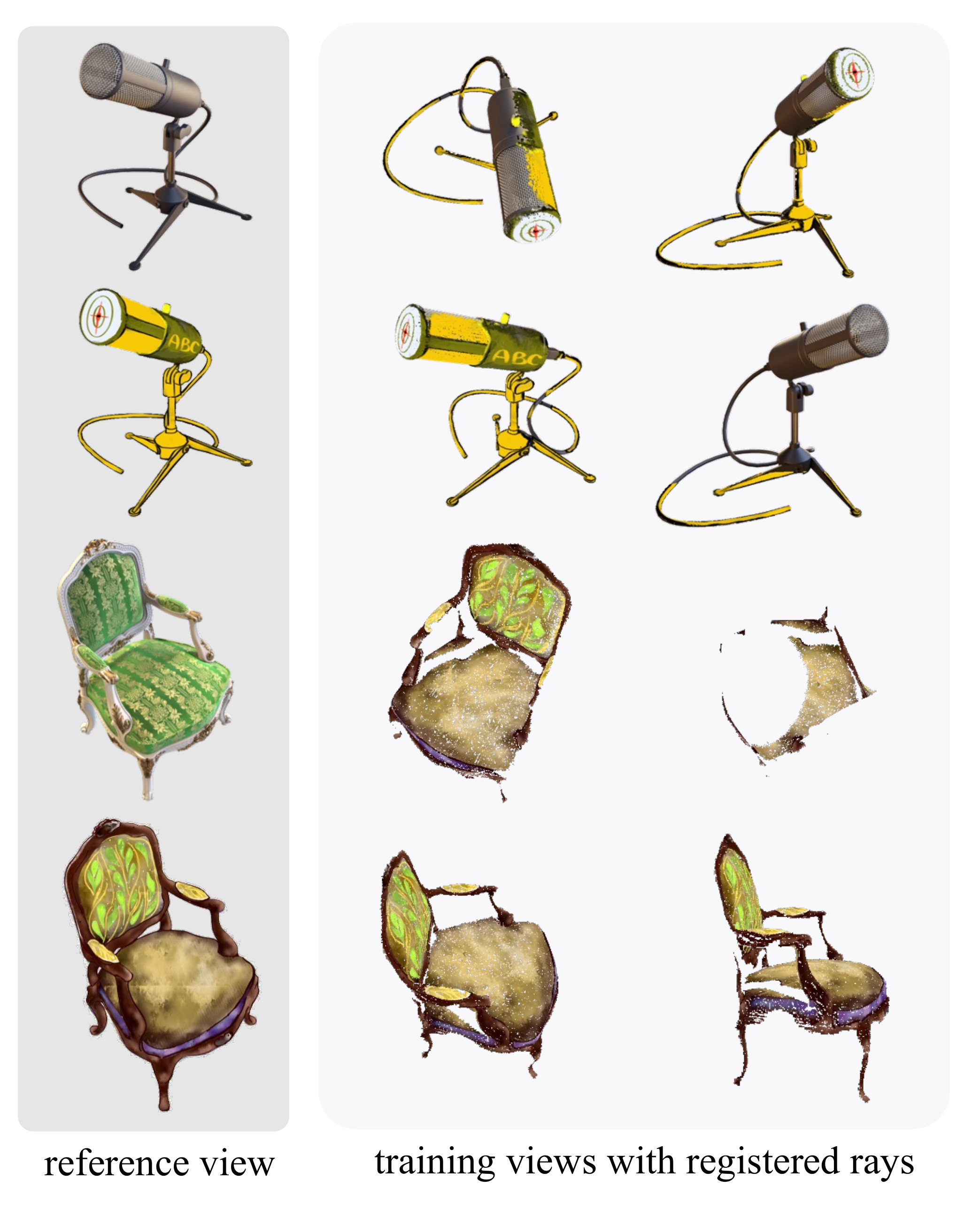}
    \caption{Two examples to visualize registered rays in R$^3$. We paste pseudo-rays on content images in the first example for a better presentation.}
    \label{fig:rayreg_vis}
\end{figure}

\paragraph{Reference ray registration.}
\cref{fig:rayreg_vis} gives two concrete examples of how ray registration provides supervision in reference-dependent areas. 
Rays related to the stylized reference $S_R$ are projected to each training view to provide pseudo-ray supervision.

\begin{figure}[ht]
    \centering
    \includegraphics[width=1.0\linewidth]{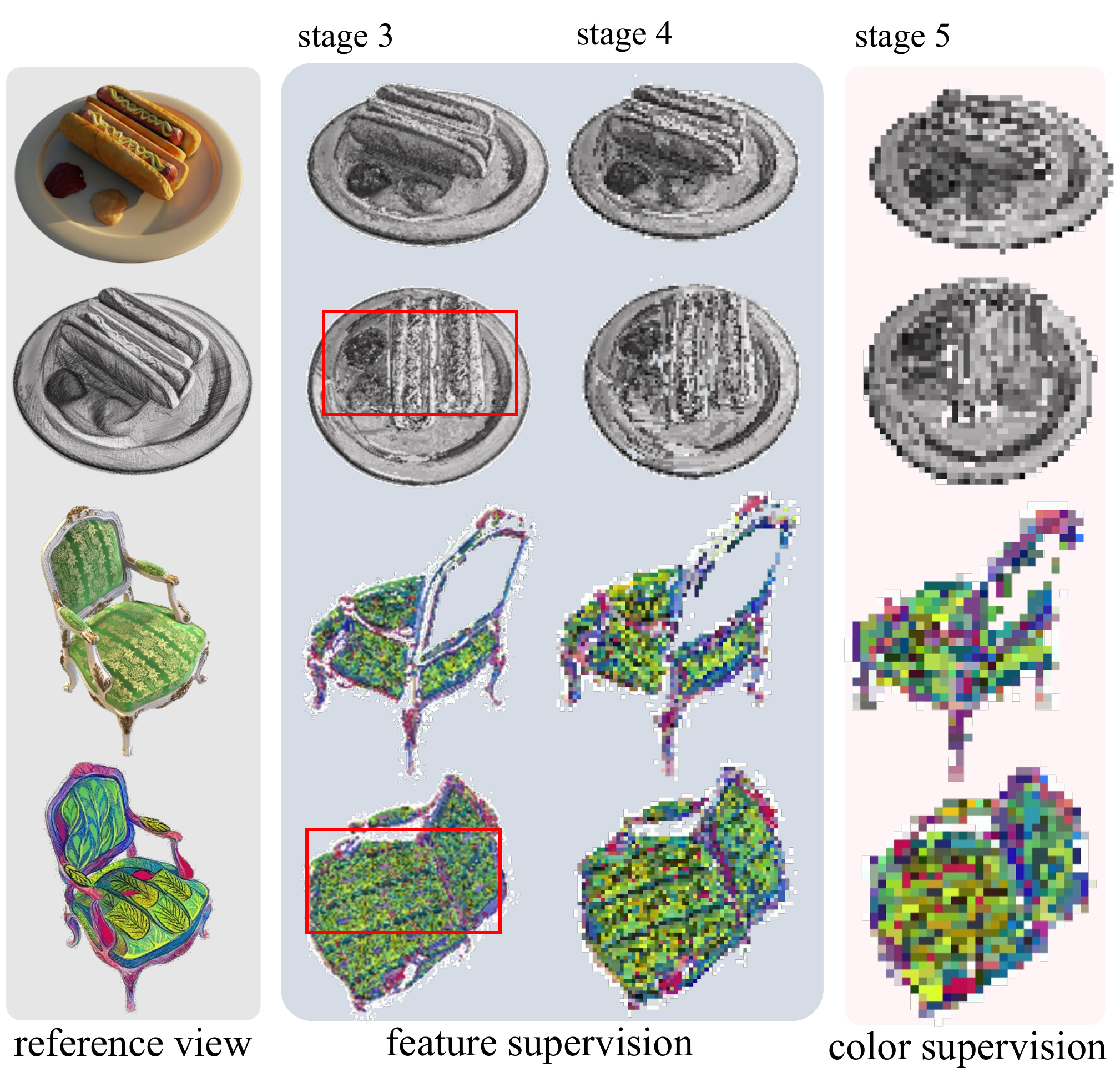}
    \caption{Two examples of patch-wise replacement on VGG feature at the last three stages to visualize the semantic correspondence. Color mismatch problems in shallow semantic features are highlighted.}
    \label{fig:replace_vis}
\end{figure}

\paragraph{Template-based feature matching.}
Except for explicit supervision in R$^3$, the implicit supervision provided by TCM is essential to occluded regions. 
\cref{fig:replace_vis} shows two examples of patch-wise replacement results.
For guidance feature $F_{G}$, we select VGG features at stages 3 and 4. 
Since the patch-wise semantic feature is a high-level representation for each patch, the receptive field is much larger than the corresponding image patch.

Conversely, directly using patch replacement results at the same stages for the color supervision $\mathcal{L}_{\mathrm{color}}$ may result in a color mismatch problem, as highlighted in~\cref{fig:rayreg_vis}.
This problem is mainly caused by the receptive field difference between the feature patch and the image patch.
Hence, as mentioned in~\cref{sec:tech_details}, we evaluate feature distances at the last VGG stage for color-matching supervision.

\paragraph{Loss balancing ablation.}
In addition to the ablation studies on the microphone example provided in Sec.~\textcolor{red}{4.4}, we conduct another ablation on the scene flower to discuss the effectiveness of color-matching loss $\mathcal{L}_{\mathrm{color}}$ and the smooth content update strategy, which is described in Sec.~\textcolor{red}{4.1}.

\begin{figure}[ht]
    \centering
    \includegraphics[width=1.0\linewidth]{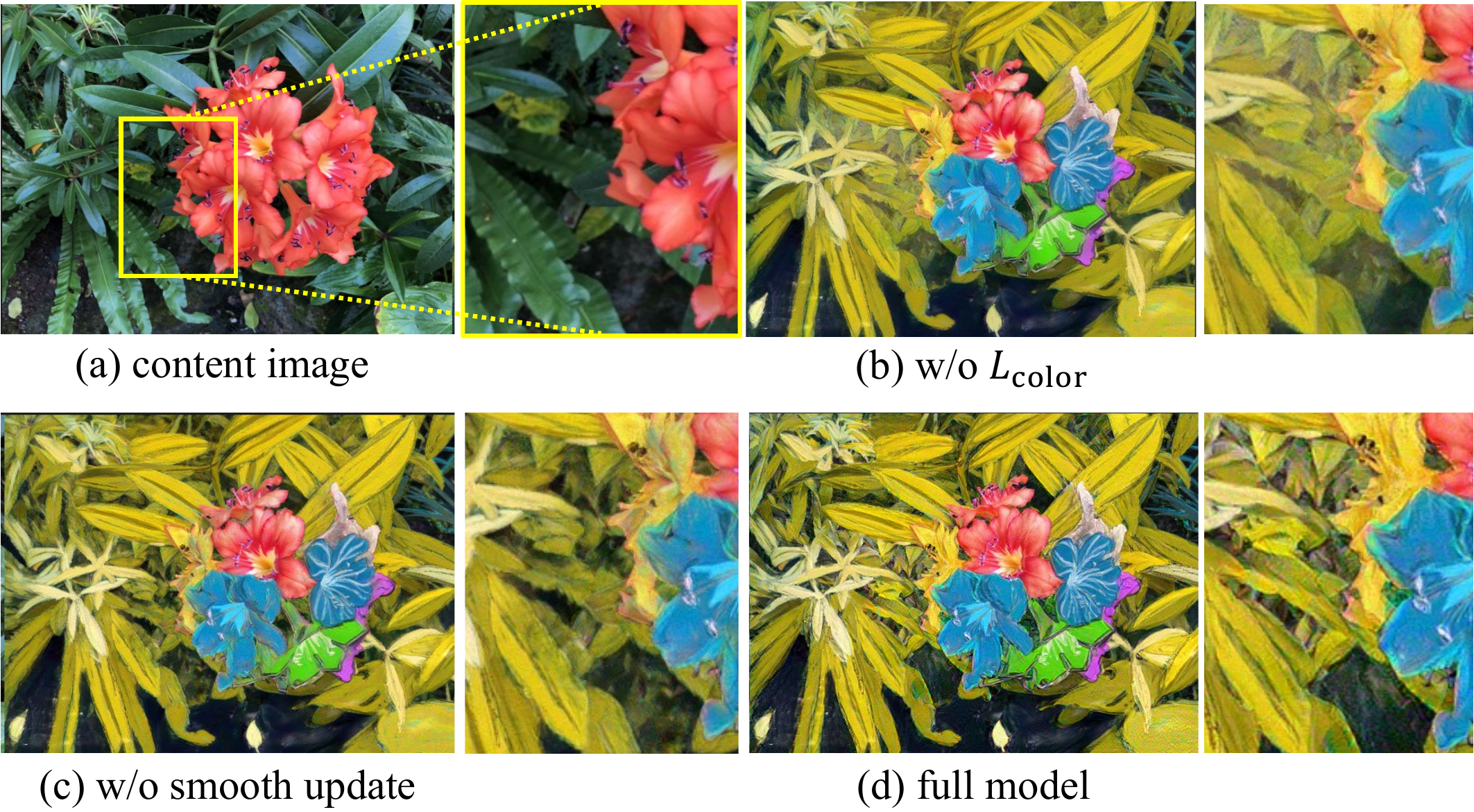}
    \caption{Ablation on the color-matching loss and smooth update strategy. 
    The occluded region is zoomed in.}
    \vspace{-2mm}
    \label{fig:ablation_v2}
\end{figure}

For the same content view in \cref{fig:ablation_v2}~(a), the color mismatch problem would exist in occluded regions when we remove the color-matching loss $\mathcal{L}_{\mathrm{color}}$, as shown in \cref{fig:ablation_v2}~(b).
In \cref{fig:ablation_v2}~(c), we find that the stylized view without applying the smooth update strategy leads to occluded regions being under-stylized, which implies that the quality of semantic correspondence in the original content domain needs to be enhanced by TCM. 
A full model in \cref{fig:ablation_v2}~(d) clearly shows a satisfying stylization result in terms of both color and style.

\begin{figure}[H]
    \centering
    \includegraphics[width=1.0\linewidth]{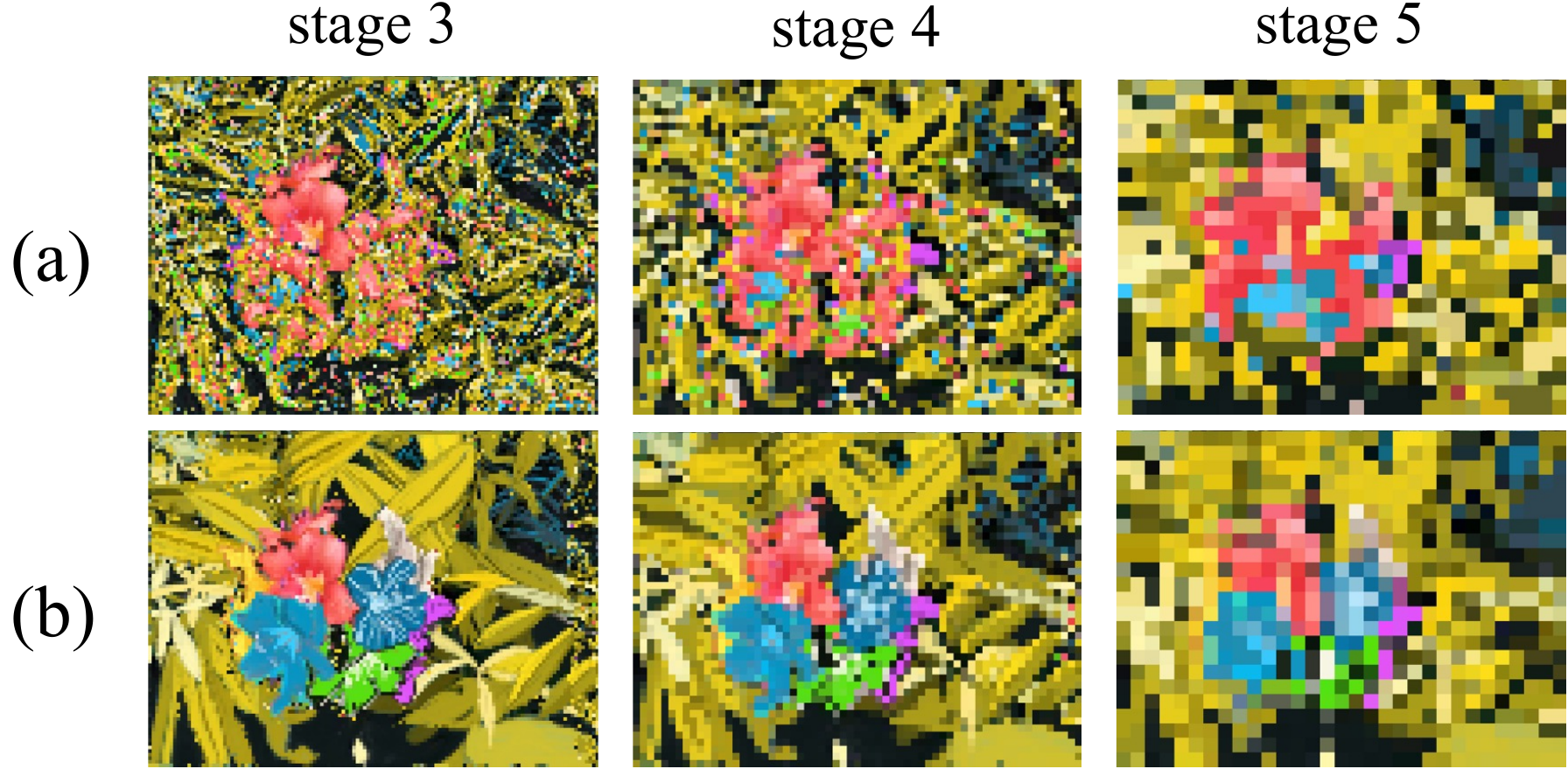}
    \caption{Patch-wise replacement results on features from the last three stages of VGG backbone. (a)~Matching with the style reference directly. (b)~Matching with the content reference~(TCM).}
    \label{fig:ablation_vis_match}
\end{figure}

\paragraph{Discussion on TCM matching. }
We also validate the how effectiveness of the patch-wise matching scheme in TCM. 
Unlike epipolar correspondence, the deep semantic feature is calculated in 2D patch-wisely. The correspondence is only computed once and costs around 2 seconds for a set with 100 images.
As shown in~\cref{fig:ablation_vis_match}~(a), a direct match with the stylized view often fails to get desired correspondence due to the domain gap in the semantic feature space.
Conversely, in~\cref{fig:ablation_vis_match}~(b), TCM matches features within the same content domain. 
Hence the semantic correspondence is preserved at each level of semantic features.

\section{More Results}
\label{sec:supp_comp}

\begin{figure}[h]
    \centering
    \href{https://youtu.be/XpIdlx2FmkA}
    {\includegraphics[width=1.0\linewidth]{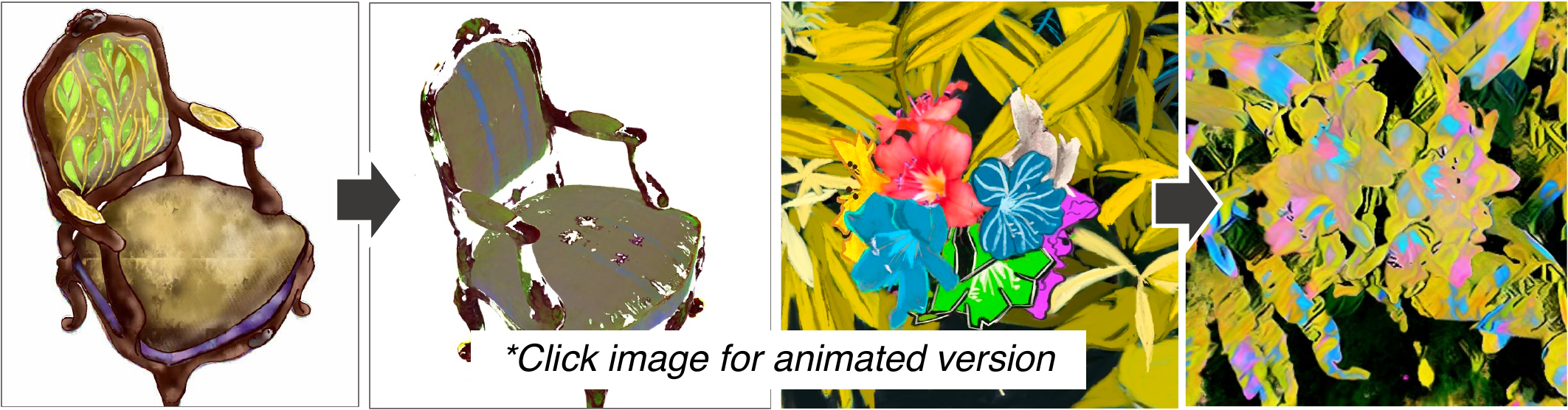}}
    \caption{Examples with INS.}
    \label{fig:eccv}
\end{figure}

\begin{figure*}[h]
    \centering
    \includegraphics[width=1.0\linewidth]{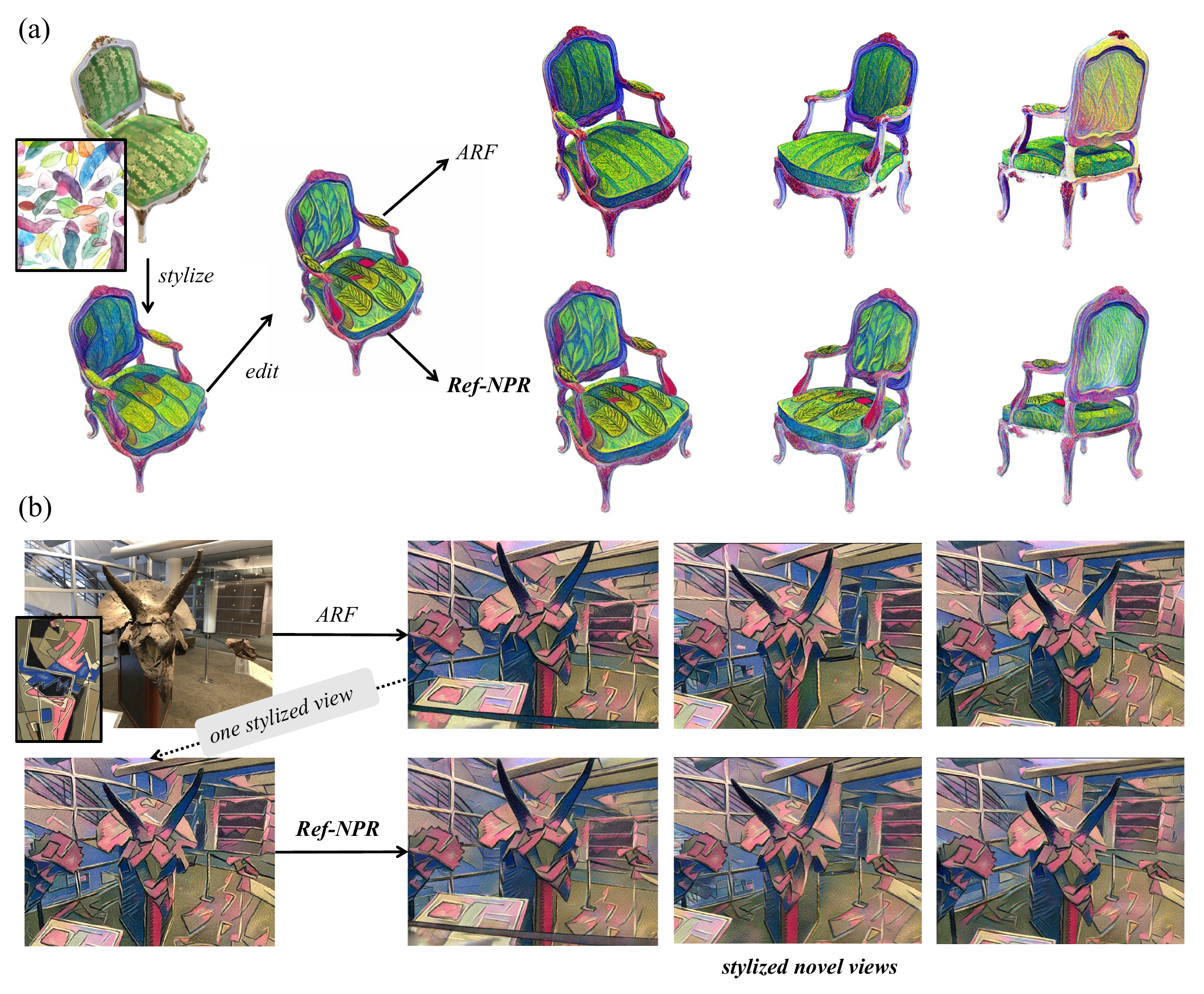}
    \caption{Examples to show the flexibility of Ref-NPR: (a) reference editing based on a stylized view, and (b) reproducing novel-view stylization given one stylized view generated by ARF~\cite{zhang2022arf} as reference.}
    \label{fig:compare_flex}
\end{figure*}

\paragraph{Comparsion with INS. } We test INS~\cite{fan2022unified} on examples with the same reference cases in Fig.~\textcolor{red}{5} and~\cref{fig:compare_flex}. Results are shown in~\cref{fig:eccv}. Due to its simple supervision design, INS cannot generate satisfying results which contain local correspondence.

\begin{figure*}[p]
    \centering
    \includegraphics[width=1.0\linewidth]{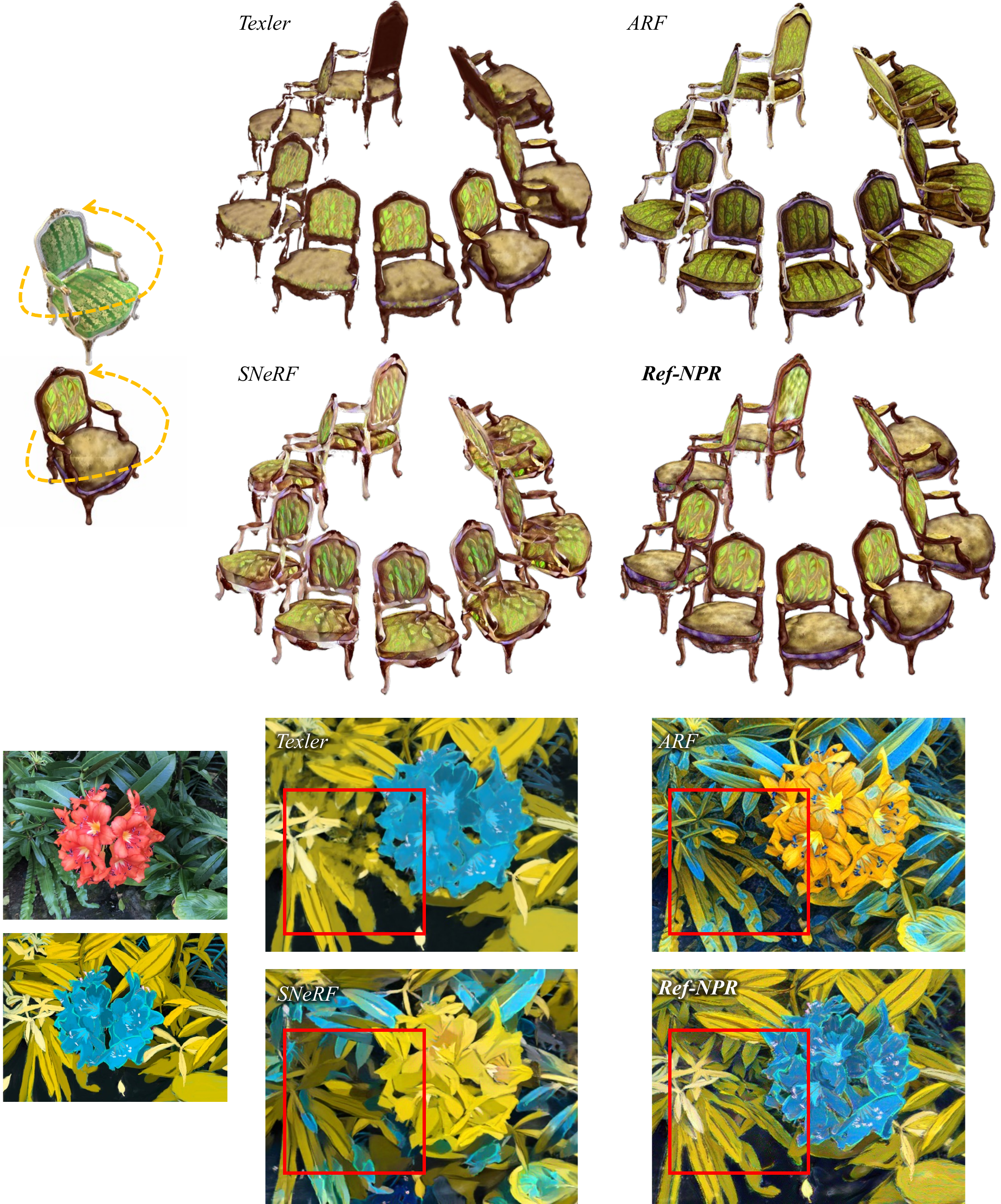}
    \caption{Additional examples for qualitative comparisons.}
    \label{fig:supp_compare}
\end{figure*}

\begin{figure*}[p]
    \centering
    \includegraphics[width=1.0\linewidth]{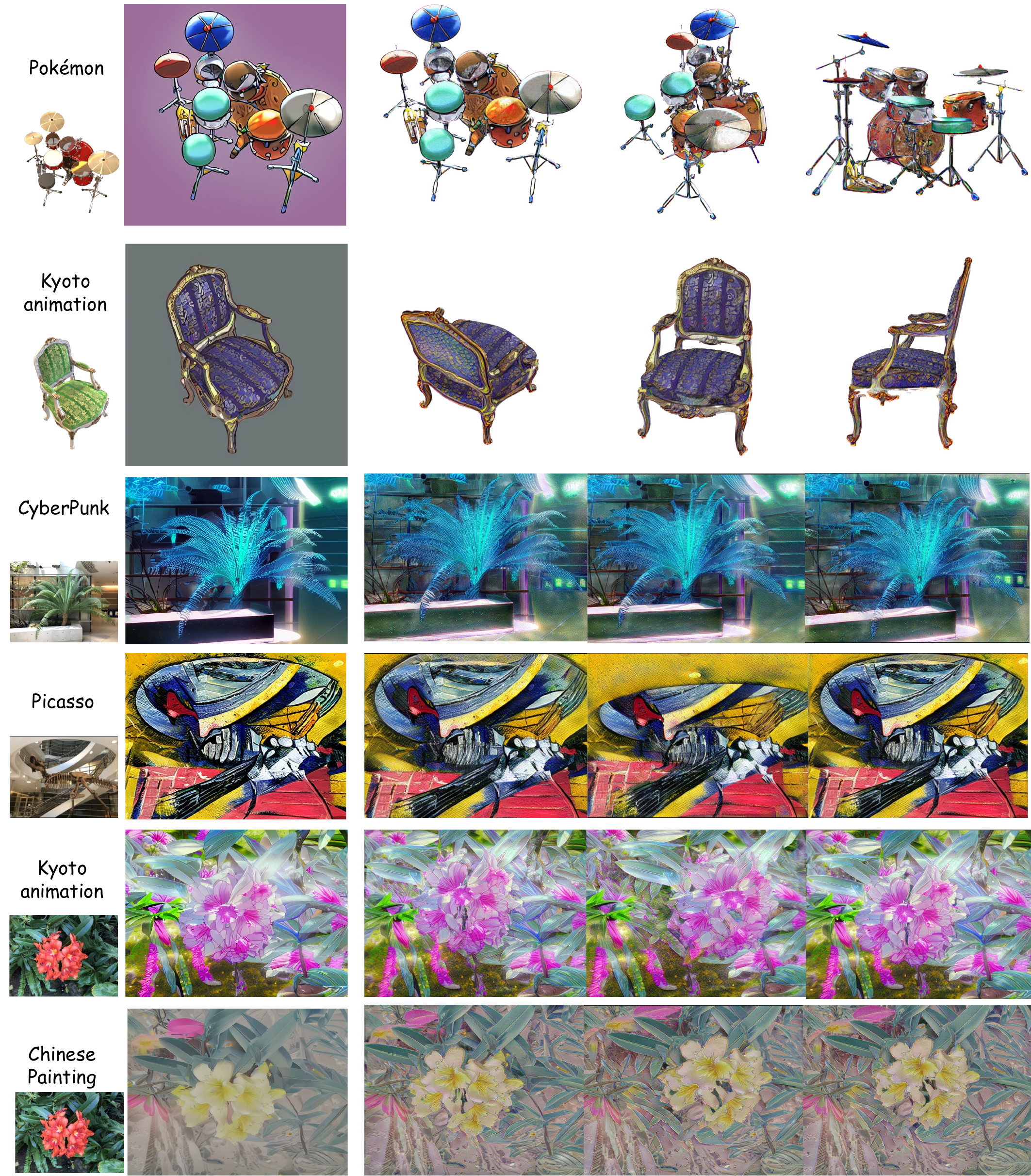}
    \caption{Controllable scene stylization with ControlNet~\cite{zhang2023adding} and Ref-NPR. A text-driven stylization with an image diffusion model is used to generate reference (the second column), then Ref-NPR can propagate it to the whole scene.}
    \label{fig:controlnet}
\end{figure*}

\begin{figure*}[p]
    \centering
    \includegraphics[width=0.94\linewidth]{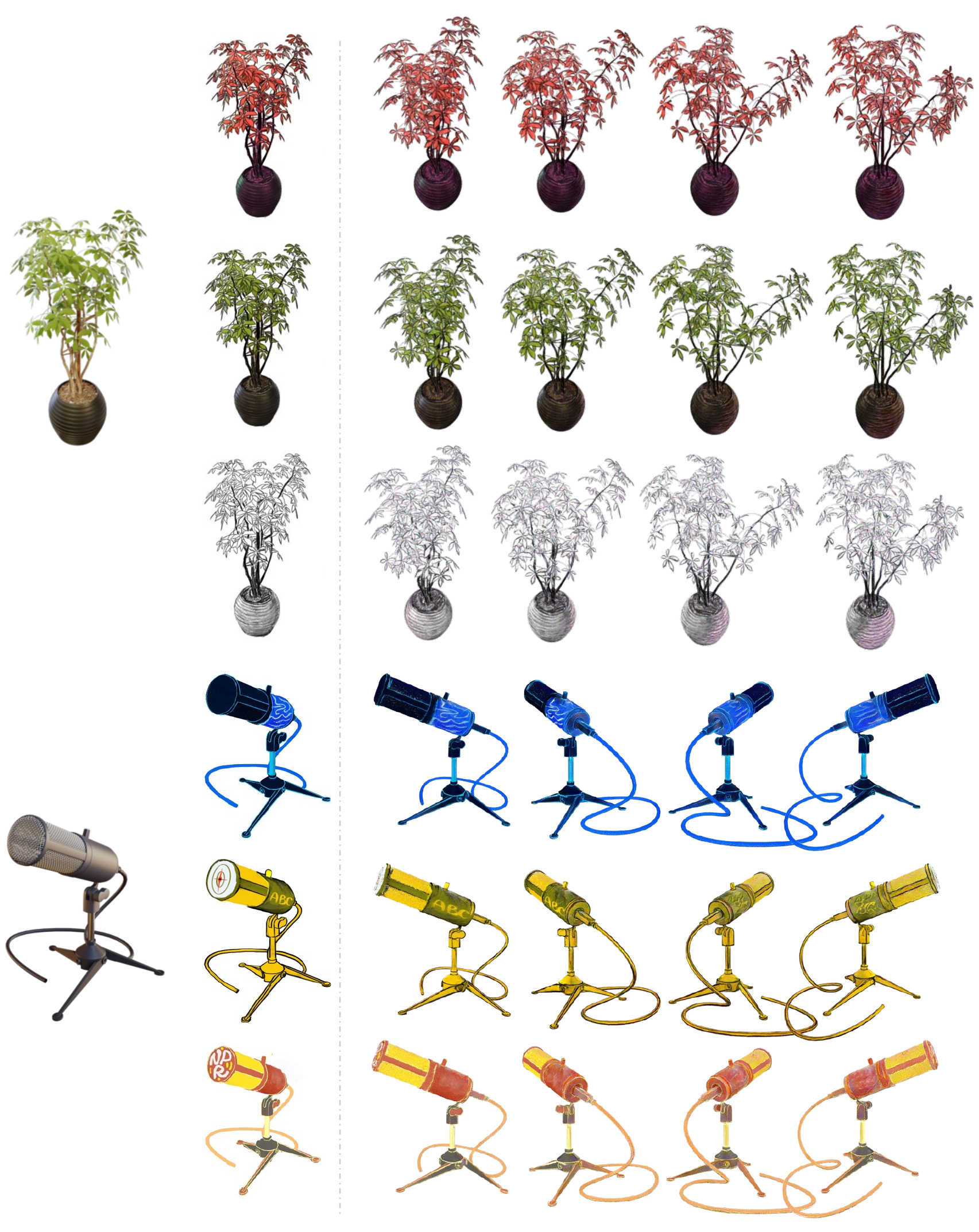}
    \caption{Examples to show the controllability of Ref-NPR with hand drawing styles. Stylized novel-view rendering results are satisfactory with references in different styles. }
    \label{fig:compare_control}
\end{figure*}

\paragraph{More comparisons.}
\cref{fig:supp_compare} offers two additional examples to compare our method with \cite{zhang2022arf, Texler20fewshot, nguyen2022snerf}.
As discussed in Sec.~\textcolor{red}{4.2}, Texler can generate novel-view stylized results with a proper color distribution, but consistent results with the reference stylized view can be only obtained under the condition that the test camera pose is around the reference.
More specifically, it fails to generate reasonable style in the occluded regions and has some flickering or ghosting artifacts in a continuous sequence.
Two scene stylization methods~\cite{zhang2022arf, nguyen2022snerf} are unable to find a desired style mapping to the entire scene. Neither in the reference-related regions nor the occluded regions.
By contrast, results generated by Ref-NPR keep both semantic correspondence and geometric consistency with the reference view.

\paragraph{Flexibility \& controllability.}
In Sec.~\textcolor{red}{5}, we show the ability of Ref-NPR to adapt with an arbitrary image as reference.
\cref{fig:compare_flex} gives two examples to demonstrate the flexibility of Ref-NPR, where the stylized reference view is generated by selecting one stylized view from ARF for each scene.
In \cref{fig:compare_flex}~(a), we manually edit the selected view and take it as the style reference.
Ref-NPR faithfully reproduces the textures in the edited regions.
Meanwhile, as shown in \cref{fig:compare_flex}~(b), our method can reproduce the original novel-view stylizations by ARF through feeding in a stylized view as reference, which requires high-quality semantic correspondence.

Except for the local editing and scene stylization reproducing, the controllability of Ref-NPR can also be represented by adapting scene stylization to various styles.
\cref{fig:compare_control} shows two examples of applying multiple styles to the same scene.
Ref-NPR is capable of producing a faithful stylization result for each style owing to the modeling of cross-view semantic correspondence.
Additionally, as shown in~\cref{fig:controlnet}, powered by controllable diffusion models~\cite{zhang2023adding}, Ref-NPR is capable of text-driven controllable scene stylization as well.

\clearpage


\end{document}